\documentclass[aps,pre,onecolumn,groupedaddress,longbibliography,superscriptaddress,showpacs,nofootinbib,notitlepage]{revtex4-2}

\usepackage{graphicx}
\usepackage{epstopdf}
\usepackage{bm}
\usepackage{amssymb}
\usepackage{psfrag}
\usepackage{color}
\usepackage{amsbsy}      
\usepackage{amsmath}
\usepackage{mathtools}
\usepackage{dsfont}
\usepackage{lineno}
\usepackage{comment}
\usepackage{hyperref}
\usepackage{hhline}
\usepackage{tcolorbox}
\usepackage{makecell}
\usepackage{booktabs}
\usepackage[algo2e]{algorithm2e}
\usepackage[noend]{algorithmic} 
\usepackage{algorithm}
\usepackage{dsfont}
\usepackage{url}
\usepackage{braket}
\usepackage{etoolbox}
\usepackage{soul, color}
\usepackage{amsopn}
\def\ie{\textit{i.e.}}
\def\eg{\textit{e.g.}}
\newcommand{\etal}{\textit{et al}.}

\begin{document}
\title{Visualizing high-dimensional loss landscapes with {Hessian} directions}

\author{Lucas B\"{o}ttcher}
\email{lbottcher@ufl.edu}
\affiliation{Department of Computational Science and Philosophy, Frankfurt School of Finance and Management, Frankfurt am Main, 60322, Germany}
\affiliation{Department of Medicine, University of Florida, Gainesville 32610, FL, USA}
\author{Gregory Wheeler}
\email{g.wheeler.de}
\affiliation{Department of Computational Science and Philosophy, Frankfurt School of Finance and Management, Frankfurt am Main, 60322, Germany}
\date{\today}
\begin{abstract}
Analyzing geometric properties of high-dimensional loss functions, such as local curvature and the existence of other optima around a certain point in loss space, can help provide a better understanding of the interplay between neural network structure, implementation attributes, and learning performance. In this work, we combine concepts from high-dimensional probability and differential geometry to study how curvature properties in lower-dimensional loss representations depend on those in the original loss space. We show that saddle points in the original space are rarely correctly identified as such in expected lower-dimensional representations if random projections are used. The principal curvature in the expected lower-dimensional representation is proportional to the mean curvature in the original loss space. Hence, the mean curvature in the original loss space determines if saddle points appear, on average, as either minima, maxima, or almost flat regions. We use the connection between expected curvature in random projections and mean curvature in the original space (\ie, the normalized Hessian trace) to compute Hutchinson-type trace estimates without calculating Hessian-vector products as in the original Hutchinson method. Because random projections are not suitable to correctly identify saddle information, we propose to study projections along dominant Hessian directions that are associated with the largest and smallest principal curvatures. We connect our findings to the ongoing debate on loss landscape flatness and generalizability. Finally, for different common image classifiers and a function approximator, we show and compare random and Hessian projections of loss landscapes with up to about $7\times 10^6$ parameters.
\end{abstract}
\maketitle
\section{Introduction}
Every deep neural network loss function depends on parameters, $\theta \in \mathbb{R}^N$, which induce a loss landscape that is typically in high dimensions~\cite{cooper2021global}. The loss landscape associated with a neural network and the performance of optimization procedures that operate on it are each influenced by several factors, including the structural properties of the neural network \cite{hornik1989multilayer,hornik1991approximation,DBLP:conf/iclr/ParkYLS21} and a range of implementation attributes one may choose \cite{hardt2016train,wilson2017marginal,choi2020on,baldassi2020shaping,pittorino2021entropic}.  The exact nature of these factors and how their combinations influence learning performance, however, remain largely unknown.

One way to better understand the interplay between neural network structure, implementation attributes, and learning performance is through a better understanding of the geometric properties of loss landscapes. For example, Keskar \etal~\cite{DBLP:conf/iclr/KeskarMNST17} analyze the local curvature around candidate minimizers via the spectrum of the underlying Hessian to characterize the flatness and sharpness of loss minima, and Dinh \etal~\cite{DBLP:conf/icml/DinhPBB17} demonstrate that reparameterizations can render flat minima sharp without affecting generalization properties. Another approach, which holds out the promise of visualizing the curvature around a point and the existence, if any, of nearby optima, aims to visualize high-dimensional loss functions by a lower-dimensional (and often random) projection of two or three dimensions~\cite{DBLP:journals/corr/GoodfellowV14,DBLP:conf/nips/Li0TSG18,DBLP:conf/emnlp/HaoDWX19,wu2020adversarial,DBLP:conf/ida/HoroiHRLWK22}. Building on this approach, Horoi \etal~\cite{DBLP:conf/ida/HoroiHRLWK22} pursue improvements in learning by dynamically sampling points in projected low-loss regions surrounding local minima during training.

Given the advantages afforded by the promise of visualizing loss landscape optima, one may ask: how do the curvature properties in low-dimensional visualizations of high-dimensional functions depend on the curvature properties of the original, high-dimensional loss space? Specifically, do random two- and three-dimensional loss projections meaningfully represent convexity and concavity properties of high-dimensional loss functions? In short, they do not. Saddle points are generally misrepresented in the expected lower-dimensional representation of an original, high dimensional loss space. Yet, due to the exponential proliferation of saddle points in high dimensions~\cite{baldi1989neural,DBLP:conf/nips/DauphinPGCGB14}, a critical point in high-dimensional loss space is almost certain to be a saddle rather than a minimum. Nevertheless, random projections {\em can} be useful to obtain curvature estimates, as we shall demonstrate by establishing a connection between expected curvature in random loss projections and Hessian trace estimates based on Hutchinson's method~\cite{hutchinson1989stochastic}.  

Our contributions in this paper are threefold. First, we explain why random projections from a high-dimensional space do not preserve curvature information. This argument is developed in Section~\ref{sec:projection_curvature}. One consequence of this result is that saddle points of an original high-dimensional loss landscape are not identified correctly in the associated expected low-dimensional random projection, a point we illustrate with simulation examples in Section~\ref{sec:examples_extracting_curvature_info}. Second, while principal curvatures associated with a random projection are not weighted ensemble means of the principal curvatures in the original high-dimensional space, the principal curvatures in a low-dimensional projection {\em are} given by functions of weighted ensemble means of the Hessian elements in the original, high-dimensional space. This result is developed in Section~\ref{sec:projection_curvature} as well, and it is also illustrated by a simulation example in Section~\ref{sec:examples_extracting_curvature_info}. A complementary simulation example in Section~\ref{sec:hessian_trace} shows how one can estimate mean curvature from the slope of random loss projections. Together, Section~\ref{sec:projection_curvature} is intended to be a self-contained description of the relationship between principal curvatures of a loss function and the principal curvatures associated with a lower-dimensional, random projection. 

Our third contribution builds on the results reported in Sections~\ref{sec:projection_curvature} and  \ref{sec:examples_extracting_curvature_info} to propose an efficient method for estimating and visualizing Hessian information in high-dimensional loss landscapes. Instead of using random projections to visualize loss functions, we propose in Sections~\ref{sec:hessian_directions} and \ref{sec:appl_nn} to analyze projections along {\em dominant Hessian directions} that are associated with the largest-magnitude positive and negative principal curvatures, \ie, those directions along which the magnitude of positive and negative changes of the loss function is largest. For different common image classifiers and a function approximator as studied in~\cite{adcock2021gap}, we show and compare random and Hessian projections of loss landscapes with up to about $7\times 10^6$ parameters. In accordance with related works unlocking Hessian information in non-linear and high-dimensional settings~\cite{yao2020pyhessian,liao2021hessian}, our proposal uses Hessian-vector products (HVPs), thereby bypassing the computational expense of computing a Hessian matrix. In short, the role of random projections in our approach is to estimate mean curvature and weighted ensemble means of elements of the original high-dimensional Hessian.

Finally, we conclude our paper in Section~\ref{sec:conclusion} with a discussion of our results. Our source code is publicly available at~\cite{GitLab}.
\section{Principal curvature in random projections}
\label{sec:projection_curvature}
To describe the connection between the principal curvature of a loss function $L(\theta)$ and that associated with a lower-dimensional, random projection, we provide in Section~\ref{sec:overview} an overview of concepts from differential geometry~\cite{lee2006riemannian,berger2012differential,kuhnel2015differential} that are useful to mathematically describe curvature in high-dimensional spaces. In Sections~\ref{sec:random_projections} and \ref{sec:principal_curvature}, we  show the relationship between principal curvature and random projections. Finally, in Section~\ref{sec:hessian_trace_estimates}, we highlight a relationship between measures of curvature presented in  Section~\ref{sec:principal_curvature} and Hutchinson-type Hessian trace estimates.

\subsection{Differential and information geometry concepts}
\label{sec:overview}
In differential geometry, the principal curvatures are the eigenvalues of the shape operator (or Weingarten map\footnote{Some authors distinguish between the shape operator and the Weingarten map depending on if the change of the underlying tangent vector is described in the original manifold or in Euclidean space (see, \eg, chapter 3 in \cite{crane2018discrete}).}), a linear endomorphism defined on the tangent space $T_p$ of $L$ at a point $p$. For the original high-dimensional space, we have $(\theta,L(\theta))\subseteq \mathbb{R}^{N+1}$ and there are $N$ principal curvatures $\kappa_1^\theta\geq \kappa_2^\theta \geq \dots \geq \kappa_N^\theta$. At a non-degenerate critical point $\theta^*$ where the gradient $\nabla_\theta L$ vanishes, the matrix of the shape parameter is given by the Hessian $H_\theta$ with elements $(H_\theta)_{ij}=\partial^2 L/(\partial \theta_i \partial \theta_j)$ ($i,j\in\{1,\dots, N\}$).\footnote{A critical point is degenerate if the Hessian $H_{\theta}$ at this point is singular (\ie, $\mathrm{det}(H_{\theta})=0$). At degenerate critical points, one cannot use the eigenvalues of $H_{\theta}$ to determine if the critical point is a minimum (positive definite $H_{\theta}$) or a maximum (negative definite $H_{\theta}$). Geometrically, at a degenerate critical point, a quadratic approximation fails to capture the local behavior of the function that one wishes to study.} Some works refer to the Hessian as the ``curvature matrix''~\cite{DBLP:journals/jmlr/Martens20} or use it to characterize the curvature properties of $L(\theta)$~\cite{DBLP:conf/nips/Li0TSG18}. In the vicinity of a non-degenerate critical point $\theta^*$, the eigenvalues of the Hessian $H_\theta$ are the principal curvatures and describe the loss function in the eigenbasis of $H_\theta$ according to
\begin{equation}
L(\theta^*+\Delta \theta)=L(\theta^*)+\frac{1}{2}\sum_{i=1}^N \kappa_i^{\theta}\Delta \theta_i^2\,.
\end{equation}
The Morse lemma states that, if a critical point $\theta^*$ of $L(\theta)$ is non-degenerate, then there exists a chart $(\tilde{\theta}_1,\dots,\tilde{\theta}_N)$ in a neighborhood of $\theta^*$ such that
\begin{equation}
\label{lem:Morse}
L(\tilde{\theta}) = -\tilde{\theta}^2_1 - \cdots - \tilde{\theta}^2_i + \tilde{\theta}^2_{i +1} + \cdots + \tilde{\theta}^2_N + L(\theta^*)\,, 
\end{equation}
where $\tilde{\theta}_i(\theta^*)=0$ for $i\in\{1,\dots,N\}$. The loss function $L(\tilde{\theta})$ in Eq.~\eqref{lem:Morse} is decreasing along $i$ directions and increasing along the remaining $i+1$ to $N$ directions. Further, the index $i$ of a critical point $\theta^*$ is the number of linearly independent decreasing dimensions of $L$ near $\theta^*$ (\ie, the number of negative eigenvalues of the Hessian $H_\theta$ at that point).  

In the standard basis, the Hessian is
\begin{equation}
H_{\theta}\coloneqq \nabla_\theta \nabla_\theta L(\theta)=\begin{pmatrix}
\frac{\partial^2 L}{\partial \theta_1^2} & \cdots & \frac{\partial^2 L}{\partial \theta_1 \partial \theta_N} \\
\vdots  & \ddots & \vdots \\
\frac{\partial^2 L}{\partial \theta_N \partial \theta_1} & \cdots & \frac{\partial^2 L}{\partial \theta_N^2}
\end{pmatrix}\,.
\label{eq:H_theta}
\end{equation}
Closely related to the Hessian is the Fisher information matrix (FIM)~\cite{casella2002statistical,lehmann2006theory}, whose elements $F_{ij}$ are given by
\begin{equation}
\label{eq:FIM_product}
 F_{ij} = \mathbb{E} \left[ \frac{\partial}{\partial\theta_i} \log p(X  \mid \theta) \frac{\partial}{\partial\theta_j} \log p(X  \mid \theta) \right]\,,
\end{equation}
where the expectation is taken over the random variable $X$ and $p(x \mid \theta)$ denotes the probability density function associated with $X$ conditioned on the parameter $\theta$. The probability that $X$ falls within the infinitesimal interval $[x,x+\mathrm{d}x]$, given a known value of $\theta$, is $p(x \mid \theta)\,\mathrm{d}x$. Provided that certain regularity conditions such as
\begin{equation}
    \frac{\mathrm{d}}{\mathrm{d}\theta_i}\mathbb{E}\left[\frac{\partial}{\partial \theta_j}\log p(X \mid \theta)\right]=\int \frac{\partial}{\partial \theta_i}\left[\left(\frac{\partial}{\partial \theta_j}\log p(x \mid \theta)\right) p(x \mid \theta)\right]\,\mathrm{d}x
\end{equation}
are satisfied~\cite{casella2002statistical,rao2017advanced}, we may rewrite Eq.~\eqref{eq:FIM_product} as
\begin{equation}
 F_{ij} = - \mathbb{E} \left[ \frac{\partial^2}{\partial\theta_i \partial \theta_j} \log p(X  \mid \theta)  \right]\,.
 \label{eq:FIM}
\end{equation}
Notice the structural similarity between $F_{ij}$ and the elements of $H_\theta$ [see Eq.~\eqref{eq:H_theta}]. In natural gradient descent \cite{amari1998natural}, the metric associated with a stochastic feedforward neural network is the FIM.

Considering a neural network with input-output pairs $(x,y)$ and parameters $\theta$, a possible choice of a model distribution, which takes on the role of the probability density in Eq.~\eqref{eq:FIM_product}, is $p(x,y|\theta)=p(y|x,\theta) q(x)$ with $p(y| x,\theta)=\exp(-\|y-f_\theta(x)\|_2^2/2)/\sqrt{2\pi}$. Here, $q(x)$ denotes the distribution of inputs $x$, $f_\theta(x)$ is the output of a neural network given input $x$, and $\|\cdot\|_2$ is the Euclidean norm~\cite{karakida2020universal}. The expected values associated with FIM elements $F_{ij}$ for $p(x,y|\theta)$ are calculated with respect to (w.r.t.) input-output pairs $(x,y)$ of the joint distribution $p(x,y|\theta)$.

For a set of input samples $x(t)$ ($t\in\{1,\dots,T\}$) and  corresponding outputs $f_{\theta}(t)\equiv f_{\theta}(x(t))$, the empirical Fisher information matrix is
\begin{equation}
F_\theta=\frac{1}{T}\sum_{t=1}^T\sum_{k=1}^C\nabla_\theta f_{\theta,k}(t) {\nabla_\theta f_{\theta,k}(t)}^\top\,,
\end{equation}
where $f_{\theta,k}$ is the $k$-th entry of the neural network output and $k\in\{1,\dots,C\}$~\cite{karakida2020universal,amari2000adaptive}. The empirical FIM converges towards the FIM, Eq.~\eqref{eq:FIM_product}, as $T\rightarrow\infty$. 

Given the mean squared loss function 
\begin{equation}
L(\theta)=\frac{1}{2T}\sum_{t=1}^T \|y(t)-f_\theta(t)\|_2^2\,,
\end{equation}
the Hessian $H_\theta$ of $L(\theta)$ and the empirical FIM $F_\theta$ are connected via
\begin{equation}
H_\theta = F_\theta - \frac{1}{T}\sum_{t=1}^T\sum_{k=1}^C (y_k(t)-f_{\theta,k}(t)) \nabla_\theta \nabla_\theta f_{\theta,k}(t)\,.
\label{eq:hessian_fim}
\end{equation}
Here, $y(t)$ denotes the desired output associated with sample $x(t)$. At the global optimum, which is usually attained for overparameterized neural networks~\cite{cooper2021global}, the difference $y_k(t)-f_{\theta,k}(t)$ vanishes for all $k,t$, and the empirical FIM is equal to the Hessian $H_\theta$~\cite{amari2000adaptive}.

We can also establish a connection between $H_\theta$ and $F_\theta$ for the cross-entropy loss $L(\theta)=- 1/T \sum_{t=1}^T \sum_{k=1}^C y_k(t)\log(f_{\theta,k}(t))$, where softmax outputs $f_{\theta,k}(t)$ are used to approximate binary class labels $y_k(t)$.\footnote{In this case, the underlying conditional probability distribution is $p(y| x,\theta)=\prod_{k=1}^C \left(f_{\theta,k}(t)\right)^{y_k}$~\cite{park2000adaptive}.} Similarly to Eq.~\eqref{eq:hessian_fim}, we obtain

\begin{equation}
    H_\theta =\frac{1}{T} \sum_{t=1}^T \sum_{k=1}^C  \frac{y_k(t)(\nabla_\theta f_{\theta,k}(t)\nabla_\theta f_{\theta,k}(t)^\top- f_{\theta,k}(t)\nabla_\theta \nabla_\theta f_{\theta,k}(t))}{f_{\theta,k}(t)^2}\,.
    \label{eq:hessian_fim_cross_entropy}
\end{equation}

Given the scenario in which $f_{\theta,k}(t)$ approaches $y_k(t)$ for all $k,t$ and using $\sum_{k=1}^{C} f_{\theta,k}(t)=1$ (\ie, $\sum_{k=1}^C \nabla_\theta \nabla_\theta f_{\theta,k}(t)=0$), we again have $H_\theta=F_\theta$ where

\begin{equation}
    F_\theta=\frac{1}{T} \sum_{t=1}^T \sum_{k=1}^C \frac{1}{f_{\theta,k}(t)} \nabla_\theta f_{\theta,k}(t)\nabla_\theta f_{\theta,k}(t)^\top
\end{equation}

is the empirical FIM associated with the cross-entropy loss~\cite{park2000adaptive}.

The neural network examples that we consider in Section~\ref{sec:appl_nn} are trained using cross-entropy loss. In the Appendix, we also consider a function approximation problem in which we train neural networks using a mean squared loss function.
\subsection{Random projections}
\label{sec:random_projections}
To graphically explore an $N$-dimensional loss function $L$ around a critical point $\theta^*$, one may wish to work in a lower-dimensional representation. For example, a two-dimensional projection of $L$ around $\theta^*$ is provided by
\begin{equation}
L(\theta^*+\alpha\eta+\beta\delta)\,,
\label{eq:app_loss_alpha_beta}
\end{equation}
where the parameters $\alpha,\beta\in \mathbb{R}$ scale the directions $\eta,\delta\in\mathbb{R}^N$. The corresponding graph representation is $(\alpha,\beta,L(\alpha,\beta))\subseteq\mathbb{R}^{3}$. 

In high-dimensional spaces, there exist vastly many more almost-orthogonal than orthogonal directions. In fact, if the dimension of our space is large enough, with high probability, random vectors will be sufficiently close to orthogonal \cite{HechtNielsen94}. Following this result, many related works~\cite{DBLP:conf/nips/Li0TSG18,wu2020adversarial,DBLP:conf/ida/HoroiHRLWK22} use random Gaussian directions with independent and identically distributed vector elements $\eta_i, \delta_i \sim\mathcal{N}(0,1)$ ($i\in\{1,\dots,N\}$).

We now turn to showing that $\eta,\delta$ are almost orthogonal using a concentration inequality for chi-squared distributed random variables. To do so, we first note that the scalar product of random Gaussian vectors $\eta,\delta$ is a sum of the difference between two chi-squared distributed random variables because
\begin{equation}
\sum_{i=1}^N \eta_i \delta_i=\sum_{i=1}^N \frac{1}{4}(\eta_i+\delta_i)^2-\frac{1}{4}(\eta_i-\delta_i)^2=\sum_{i=1}^N \frac{1}{2} X_i^2-\frac{1}{2} Y_i^2\,,
\label{eq:scalar_product_eta_delta}
\end{equation}
where $X_i,Y_i\sim \mathcal{N}(0,1)$. In the last step of Eq.~\eqref{eq:scalar_product_eta_delta}, we rescaled $X_i$ and $Y_i$ by a factor $\sqrt{2}$ to obtain normally distributed random variables with variance 2. Since $\mathds{E}[Z]=N$ for $Z=\sum_{i=1}^N X_i^2$, the right-hand side of Eq.~\eqref{eq:scalar_product_eta_delta} vanishes in expectation. That is, 
\begin{equation}
\lim_{N\rightarrow\infty} \frac{1}{N}\sum_{i=1}^N \eta_i \delta_i = 0\,.
\end{equation}
Hence, the vectors $\eta,\delta$ are orthogonal in the limit $N\rightarrow\infty$. For finite $N$, we can bound Eq.~\eqref{eq:scalar_product_eta_delta} using the concentration inequalities
\begin{align}
\Pr\left(\frac{1}{N}\sum_{i=1}^N X_i^2-1 \geq \epsilon +\frac{1}{2} \epsilon^2\right) &\leq e^{-\frac{N \epsilon^2}{4}}\label{eq:concentration_bound_11}
\\
\Pr\left(1-\frac{1}{N}\sum_{i=1}^N X_i^2 \geq \epsilon\right) &\leq e^{-\frac{N \epsilon^2}{4}}
\label{eq:concentration_bound_12}
\end{align}
for any $\epsilon>0$~\cite{laurent2000adaptive}. In the limit $\epsilon\rightarrow 0$, the above inequalities can be cast in the following form
\begin{align}
\Pr\left(\left|\frac{1}{N}\sum_{i=1}^N X_i^2-1\right| \geq \epsilon \right) &\leq 2 e^{-\frac{N \epsilon^2}{4}}\,.
\label{eq:concentration_bound_2}
\end{align}
Using $|N^{-1} \sum_{i} \eta_i \delta_i|=|(2 N)^{-1} \sum_i X_i^2-1+1-Y_i^2|$, we obtain
\begin{equation}
\Pr\left(\left| \frac{1}{N}\sum_{i=1}^N \eta_i \delta_i \right|\geq \epsilon \right) \leq \sqrt{2} e^{-{N \epsilon^2 \over 2}}
\label{eq:concentration}
\end{equation}
in the limit of large $N$. For further details on the derivation of Eq.~\eqref{eq:concentration}, see Appendix~\ref{app:concentration}.

\subsection{Principal curvature}
\label{sec:principal_curvature}
\begin{table}
\centering
\renewcommand*{\arraystretch}{2}
\begin{tabular}{c c}\toprule
\,\,\,Symbol\,\,\, & \,\,\,Definition\,\,\, \\ \hline
\,\,\,$H_\theta\in\mathbb{R}^{N\times N}$\,\,\, & \,\,\,\makecell[l]{Hessian in the original loss space}\,\,\, \\ \hline
\,\,\,$\kappa_i^{\theta}\in\mathbb{R}$\,\,\, & \,\,\,\makecell[l]{principal curvatures in the original loss space, \\
where $i\in\{1,\dots,N\}$ (\ie, the eigenvalues of $H_{\theta}$)}\,\,\, \\ \hline
\,\,\,$H_{\alpha,\beta}\in\mathbb{R}^{2\times 2}$\,\,\, & \,\,\,\makecell[l]{Hessian in a two-dimensional projection of an \\
$N$-dimensional loss function}\,\,\, \\ \hline
\,\,\,$\kappa_{\pm}^{\alpha,\beta}\in\mathbb{R}$\,\,\, & \,\,\,\makecell[l]{principal curvatures in a two-dimensional \\
loss projection (\ie, the eigenvalues of $H_{\alpha,\beta}$)}\,\,\, \\[4pt] \hline
\,\,\,$\bar{\kappa}^{\alpha,\beta}\in\mathbb{R}$\,\,\, & \,\,\,\makecell[l]{principal curvature in the expected, two-dimensional \\ 
loss projection (\ie, the eigenvalue of $\mathbb{E}[H_{\alpha,\beta}]$)}\,\,\, \\[4pt] \hline
\,\,\,$H\in\mathbb{R}$\,\,\, & \,\,\,\makecell[l]{mean curvature (\ie, $\sum_{i=1}^N \kappa_i^\theta/N=\bar{\kappa}^{\alpha,\beta}/N$)}\,\,\, \\ \bottomrule
\end{tabular}
\vspace{1mm}
\caption{\textbf{Overview of several quantities that are useful to characterize curvature of high-dimensional loss functions.} We summarize the employed definitions of Hessians and several derived curvature measures that are useful to quantify curvature of $N$-dimensional loss functions ${L(\theta^*)\colon \mathbb{R}^N\rightarrow\mathbb{R}}$ and their two-dimensional random projections $L(\theta^*+\alpha\eta+\beta\delta)$ at a non-degenerate critical point $\theta^*$. The vectors $\eta,\delta\in\mathbb{R}^N$ are the directions along which we project the original loss function and $\alpha,\beta\in\mathbb{R}$ are the corresponding scale factors. Random loss projections are usually based on Gaussian vectors $\eta,\delta$, which are almost orthogonal in high-dimensional spaces (\ie, for large $N$).}
\label{tab:curvature_measures}
\end{table}

With the form of random Gaussian projections in hand, we now analyze the principal curvatures in both the original and lower-dimensional spaces. The Hessian associated with the two-dimensional loss projection \eqref{eq:app_loss_alpha_beta} is
\begin{align}
\begin{split}
H_{\alpha,\beta}&=\begin{pmatrix}
\frac{\partial^2 L}{\partial \alpha^2} & \frac{\partial^2 L}{\partial \alpha \partial \beta} \\
 \frac{\partial^2 L}{\partial \beta \partial \alpha}  & \frac{\partial^2 L}{\partial \beta^2} \\
\end{pmatrix}=
\begin{pmatrix}
\sum_{i,j} \eta_i \eta_j \frac{\partial^2 L}{\partial \theta_i \theta_j} & \sum_{i,j} \eta_i \delta_j \frac{\partial^2 L}{\partial \theta_i \theta_j} \\
\sum_{i,j} \eta_i \delta_j \frac{\partial^2 L}{\partial \theta_i \theta_j}  & \sum_{i,j} \delta_i \delta_j \frac{\partial^2 L}{\partial \theta_i \theta_j} \\
\end{pmatrix}\\
&=\begin{pmatrix}
(H_\theta)_{ij} \eta^i \eta^j & (H_\theta)_{ij} \eta^i \delta^j\\
(H_\theta)_{ij} \eta^i \delta^j & (H_\theta)_{ij} \delta^i \delta^j\\
\end{pmatrix}\,,
\end{split}
\label{eq:hessian_2d}
\end{align}
where we use Einstein notation in the last equality.

In analogy with Eq.~\eqref{eq:hessian_fim}, the dimension-reduced Hessian $H_{\alpha,\beta}$ is connected to the corresponding FIM $F_{\alpha,\beta}$ according to
\begin{equation}
H_{\alpha,\beta} = F_{\alpha,\beta}-\frac{1}{T}\sum_{t=1}^T\sum_{k=1}^C (y_k(t)-f_{(\alpha,\beta),k}(t)) \nabla_{(\alpha,\beta)} \nabla_{(\alpha,\beta)} f_{(\alpha,\beta),k}(t)\,,
\label{eq:hessian_fim_alpha_beta}
\end{equation}
where $f_{(\alpha,\beta),k}(t)$ is the $k$-th entry of the output of a neural network with parameters $\theta^*+\alpha\eta+\beta\delta$. The operator $\nabla_{(\alpha,\beta)}$ denotes the gradient w.r.t.\ parameters $\alpha,\beta$. Recall that Eq.~\eqref{eq:hessian_fim} [and hence Eq.~\eqref{eq:hessian_fim_alpha_beta}] is based on a mean squared loss function. One may also use Eq.~\eqref{eq:hessian_fim_cross_entropy} to obtain a connection between the dimension-reduced Hessian and the corresponding cross-entropy FIM.

Lower-dimensional approximations $H_{\alpha,\beta}$ of the Hessian $H_\theta$ in the original high-dimensional space might be useful in generalized Gauss--Newton algorithms that are based on positive semidefinite and computationally affordable Hessian approximations~\cite{schraudolph2002fast,DBLP:conf/cdc/DiehlM19}. 

Because the elements of $\delta,\eta$ arising in $H_{\alpha,\beta}$ are distributed according to a standard normal distribution, the second derivatives of the loss function $L$ in Eq.~\eqref{eq:hessian_2d} have prefactors that are products of standard normal variables and, hence, can be expressed as sums of chi-squared distributed random variables as in Eq.~\eqref{eq:scalar_product_eta_delta}. To summarize, elements of $H_{\alpha,\beta}$ are sums of second derivatives of $L$ in the original space weighted with chi-squared distributed prefactors. 

The principal curvatures $\kappa_{\pm}^{\alpha,\beta}$ (\ie, the eigenvalues of $H_{\alpha,\beta}$) are 
\begin{equation}
\kappa_{\pm}^{\alpha,\beta}=\frac{1}{2}\left(A+C\pm \sqrt{4 B^2+(A-C)^2}\right)\,,
\label{eq:kapp_alpha_beta}
\end{equation}
where $A=(H_\theta)_{ij}\eta^i \eta^j $, $B=(H_\theta)_{ij} \eta^i \delta^j$, and $C=(H_\theta)_{ij} \delta^i \delta^j $. To the best of our knowledge, a closed, analytic expression for the distribution of the quantities $A,B,C$ is not yet known~\cite{davies1980algorithm,laurent2000adaptive,bausch2013efficient,chen2019numerical}.

We now turn to the relationship between random projections and principal curvature. The appeal of random projections is that pairwise distances between points in a high-dimensional space can be nearly preserved by a lower-dimensional linear embedding, affording a low-dimensional representation of mean and variance information with minimal distortion \cite{Matousek:2008}. The question is whether random projections also preserve curvature information, and if so, what is the nature of the relationship between random Gaussian directions and principal curvature. For instance, Li \etal~\cite{DBLP:conf/nips/Li0TSG18} assert that ``the principal curvatures of a dimensionality-reduced plot (with random Gaussian directions) are weighted averages of the principal curvatures of the full-dimensional surface''. However, our results show that the principal curvatures $\kappa_{\pm}^{\alpha,\beta}$ in a two-dimensional loss projection are weighted averages of the Hessian elements $(H_\theta)_{ij}$ in the original space and not weighted averages of the principal curvatures $\kappa_i^\theta$, which instead are solutions of an $N$-th degree polynomial. Similar arguments apply to projections with dimension larger than 2. 

In Section~\ref{sec:examples} we discuss examples that show that lower-dimensional projections of $L(\theta)$ can be misleading, since high-dimensional saddle points may appear to be minima, maxima, or almost flat regions depending on the index of the underlying Hessian $H_\theta$ in the original space.

Returning to principal curvature, since $\sum_{i,j} a_{ij} \eta^{i} \eta^{j}=\sum_{i} a_{ii} \eta^{i} \eta^{i}+\sum_{i\neq j} a_{ij} \eta^i \eta^j$ ($a_{ij}\in\mathbb{R}$), we find that $\mathds{E}[A]=\mathds{E}[C]={(H_\theta)^i}_i$ and $\mathds{E}[B]=0$ where ${(H_\theta)^i}_i\equiv \mathrm{tr}(H_\theta)=\sum_{i=1}^N \kappa_i^\theta$.\footnote{The expected values of the quantities $A$, $B$, and $C$ correspond to ensemble means \eqref{eq:ensemble_average} over $S$ realizations of random projections in the limit $S\rightarrow\infty$.} To show that the expected values of $a_{ij}\eta^i\eta^j$ ($i\neq j$) or $a_{ij}\eta^i\delta^j$ vanish, one can either invoke independence of $\eta^i,\eta^j$ ($i\neq j$) and $\eta^i,\delta^j$ or transform both products into corresponding differences of two chi-squared random variables with the same mean [see Eq.~\eqref{eq:scalar_product_eta_delta}].

Hence, the expected, dimension-reduced Hessian \eqref{eq:hessian_2d} is
\begin{equation}
\mathds{E}[H_{\alpha,\beta}]=\begin{pmatrix} {(H_\theta)^i}_i& 0\\
0 & {(H_\theta)^i}_i\\
\end{pmatrix}\,.
\label{eq:expected_hessian}
\end{equation}
The corresponding eigenvalue (or principal curvature) $\bar{\kappa}^{\alpha,\beta}$ is therefore given by the sum over all principal curvatures in the original space (\ie, $\bar{\kappa}^{\alpha,\beta}=\sum_{i=1}^N \kappa_i^\theta$). Hence, the value of the principal curvature $\bar{\kappa}^{\alpha,\beta}$ in the expected dimension-reduced space will be either positive (if the positive principal curvatures in the original space dominate), negative (if the negative principal curvatures in the original space dominate), or close to zero (if positive and negative principal curvatures in the original space cancel out each other). As a result, saddle points will not appear as such in the expected random projection.

In addition to the connection between $\bar{\kappa}^{\alpha,\beta}$ and the principal curvatures $\kappa_i^\theta$, we now provide an overview of additional mathematical relations between different curvature measures that are useful to quantify curvature properties of high-dimensional loss functions and their two-dimensional random projections.

Invoking Eq.~\eqref{eq:kapp_alpha_beta}, we can relate $\bar{\kappa}^{\alpha,\beta}$ to $\mathrm{tr}(H_\theta)$ and $\kappa_{\pm}^{\alpha,\beta}$. Because $\kappa_{+}^{\alpha,\beta}+\kappa_{-}^{\alpha,\beta}=A+C$, we have
\begin{equation}
\mathrm{tr}(H_\theta)=\bar{\kappa}^{\alpha,\beta}=\sum_{i=1}^N \kappa_i^\theta=\frac{1}{2}\left(\mathbb{E}[\kappa_{+}^{\alpha,\beta}]+\mathbb{E}[\kappa_{-}^{\alpha,\beta}]\right)\,.
\label{eq:trace_estimate}
\end{equation}
The mean curvature $H$ in the original space is related to $\bar{\kappa}^{\alpha,\beta}$ via
\begin{equation}
H=\frac{1}{N}\bar{\kappa}^{\alpha,\beta}=\frac{1}{N}\sum_{i=1}^N \kappa_i^\theta\,.
\label{eq:mean_curvature}
\end{equation}
We summarize the definitions of the employed Hessians and curvature measures in Tab.~\ref{tab:curvature_measures}.

\subsection{Hessian trace estimates}
\label{sec:hessian_trace_estimates}

Finally, we point to a connection between the above curvature measures and existing Hessian trace estimates. The trace (or unnormalized mean curvature) of the Hessian $H_\theta$ has already found applications in characterizing loss landscapes of neural networks~\cite{Mahoney:2018,yao2020pyhessian}. A common way of estimating $\mathrm{tr}(H_\theta)$ without explicitly computing all eigenvalues of $H_\theta$ is based on Hutchinson's method~\cite{hutchinson1989stochastic} and random numerical linear algebra~\cite{bai1996some,avron2011randomized}. The basic idea behind this approach is to (i) use a random vector $z\in\mathbb{R}^N$ with elements $z_i$ that are distributed according to a distribution function with zero mean and unit variance (\eg, a Rademacher distribution with $\Pr\left(z_i=\pm 1\right)=1/2$), and (ii) compute $z^\top H_\theta z$, an unbiased estimator of $\mathrm{tr}(H_\theta)$, using Hessian-vector products. That is, 
\begin{equation}
\mathrm{tr}(H_\theta)=\mathbb{E}[z^\top H_\theta z]\,.
\label{eq:hutchinson}
\end{equation}
Recall that Eq.~\eqref{eq:trace_estimate} shows that the principal curvature of the expected random loss projection, $\bar{\kappa}^{\alpha,\beta}$, is equal to $\mathrm{tr}(H_\theta)$. Instead of estimating $\mathrm{tr}(H_\theta)$ using the original Hutchinson's method \eqref{eq:hutchinson}, an alternative Hutchinson-type estimate is provided by the mean of the expected values of $\kappa_{-}^{\alpha,\beta}$ and $\kappa_{+}^{\alpha,\beta}$ [see Eq.~\eqref{eq:trace_estimate}].
\section{Illustrative examples}
\label{sec:examples}

In this section, we summarize the main concepts derived in  Section~\ref{sec:projection_curvature} in terms of different examples. These examples are intended to aid intuition on how (i) principal and mean curvatures of the original loss function relate to those of random projections, (ii) curvature-based Hessian trace estimates relate to those obtained with the original Hutchinson's method, and (iii) Hessian directions (\ie, the eigenbasis of the Hessian $H_\theta$ of $L(\theta)$) can be used for a guided visual analysis of saddle information in high-dimensional spaces.
\subsection{Extracting curvature information}
\label{sec:examples_extracting_curvature_info}
We now study two examples that will help build intuition for the concepts discussed in the previous sections. In the first example, we study a critical point $\theta^*$ of an $N$-dimensional loss function $L(\theta)$ for which (i) all principal curvatures have the same magnitude and (ii) the number of positive curvature directions is equal to the number of negative curvature directions. The mean curvature of this saddle point is zero. In accordance with Eq.~\eqref{eq:mean_curvature}, we will show that the mean curvature can be approximated by a function of ensemble means of elements of the dimension-reduced Hessian $H_{\alpha,\beta}$, and that the principal curvatures $\kappa_{\pm}^{\alpha,\beta}$ are able to correctly identify the saddle point in the majority of simulated projections. In the second example, we use a loss function associated with an unequal number of negative and positive curvature directions. For the different curvature measures derived in Section~\ref{sec:principal_curvature}, we find that random projections cannot identify the underlying saddle point.
\begin{figure}
    \centering
    \includegraphics[width=\textwidth]{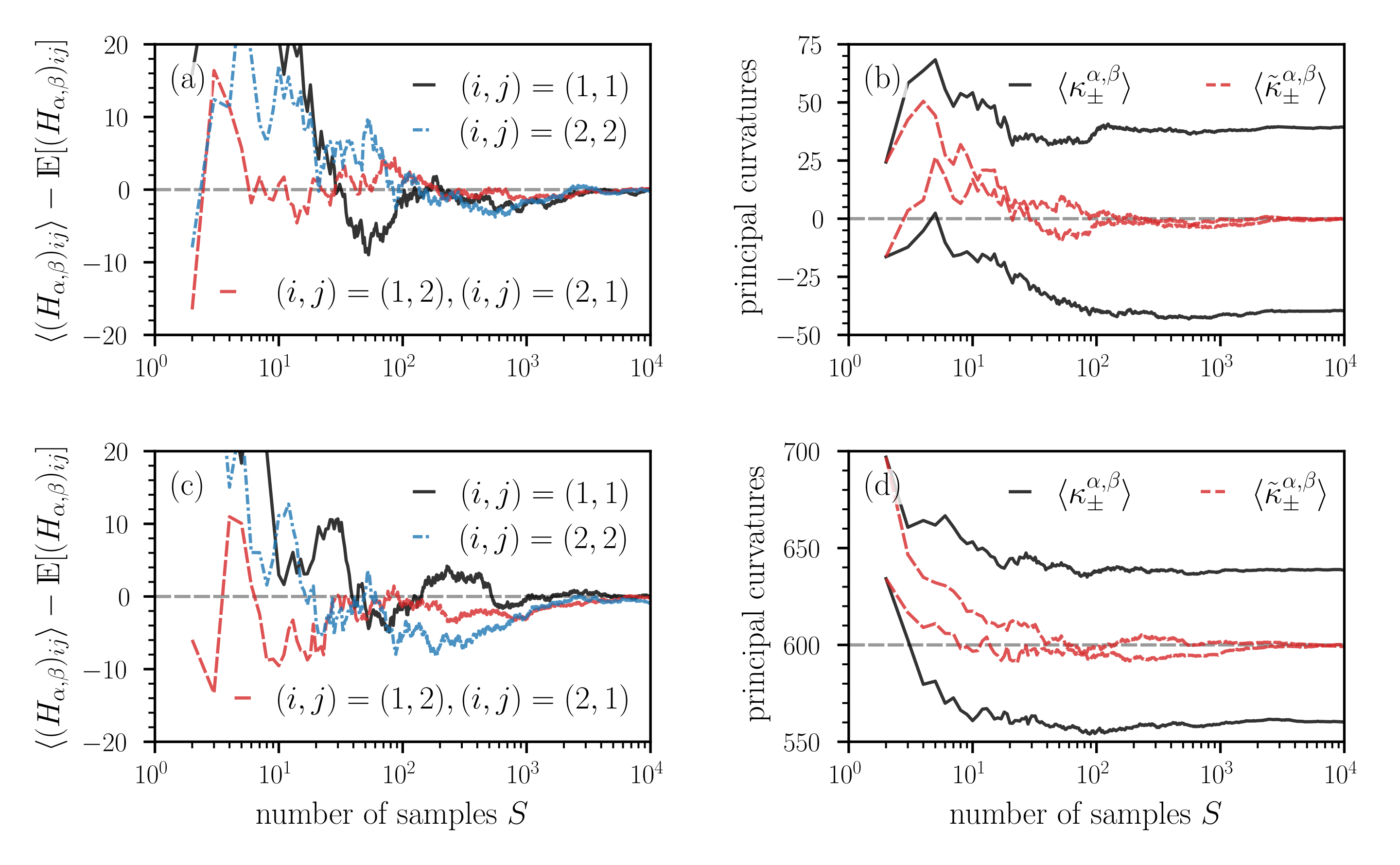}
    \caption{Convergence of the ensemble mean \eqref{eq:ensemble_average} of Hessian elements and curvatures measures as a function of the number of random projections $S$. (a,c) The deviation of the ensemble means $\langle (H_{\alpha,\beta})_{ij}\rangle$ (${i,j\in\{1,2\}}$) of Hessian elements from the corresponding expected values as a function of $S$. Notice that the expected value of the diagonal elements $(H_{\alpha,\beta})_{11}$ and $(H_{\alpha,\beta})_{22}$ is equal to $\bar{\kappa}^{\alpha,\beta}$ (\ie, to the sum of principal curvatures in the original space) [see Eqs.~\eqref{eq:expected_hessian} and \eqref{eq:trace_estimate}]. A relatively large number of random projections between $10^3$ and $10^4$ is required to keep the deviations at values smaller than about 2--4. (b,d) The ensemble means $\langle \kappa^{\alpha,\beta}_{\pm}\rangle$ [see Eq.~\eqref{eq:kapp_alpha_beta}] and $\langle \tilde{\kappa}^{\alpha,\beta}_{\pm}\rangle$ [see Eq.~\eqref{eq:kappa_tilde_sample}] as a function of $S$. Dashed grey lines represent $\bar{\kappa}^{\alpha,\beta}=\mathrm{tr}(H_\theta)$. In panels (a,b) and (c,d), the $N$-dimensional loss functions are given by Eqs.~\eqref{eq:loss_symmetric} and \eqref{eq:loss_asymmetric}, respectively. We evaluate the corresponding Hessians \eqref{eq:emp_hessian_1} and \eqref{eq:emp_hessian_2} at the saddle point $\theta^*=(\theta^*_1,\dots,\theta^*_{2n},\theta^*_{2n+1})=(0,\dots,0,1)$. In both loss functions, we set $n=500$ and in loss function~\eqref{eq:loss_asymmetric} we set $\tilde{n}=800$.}
    \label{fig:hessians_random}
\end{figure}

The loss function of our first example is
\begin{equation}
L(\theta)=\frac{1}{2}\theta_{2n+1}\left(\sum_{i=1}^n\theta_i^2-\theta_{i+n}^2\right)\,,\quad n\in\mathbb{Z}_{+}\,,
\label{eq:loss_symmetric}
\end{equation}
where we set $N=2n+1$. A critical point $\theta^*$ of the loss function \eqref{eq:loss_symmetric} satisfies
\begin{equation}
    (\nabla_\theta L)(\theta^*)=
    \left(
    \begin{array}{c}
    \theta^*_1 \theta^*_{2n+1}\\
    \vdots\\
    \theta^*_{n}\theta^*_{2n+1}\\
    -\theta^*_{n+1}\theta^*_{2n+1}\\
    \vdots\\
    -\theta^*_{2n}\theta^*_{2n+1}\\
    \frac{1}{2}\left(\sum_{i=1}^n{\theta_{i}^{{*\textsuperscript{2}}}}-{\theta_{i+n}^{{*\textsuperscript{2}}}}\right)
    \end{array}
    \right)=0\,.
\end{equation}
The Hessian at the critical point $\theta^*=(\theta^*_1,\dots,\theta^*_{2n},\theta^*_{2n+1})=(0,\dots,0,1)$ is 
\begin{equation}
H_\theta=\mathrm{diag}(\underbrace{1,\dots,1}_{n~\mathrm{times}},\underbrace{-1,\dots,-1}_{n~\mathrm{times}},0)\,.
\label{eq:emp_hessian_1}
\end{equation}
Because $H_\theta$ has positive and negative eigenvalues, the critical point is a saddle. The corresponding principal curvatures are $\kappa_i^\theta\in\{-1,1\}$ ($i\in\{1,\dots,N-1\}$) and $\kappa^\theta_N=0$. In this example, the mean curvature $H$, as defined in Eq.~\eqref{eq:mean_curvature}, is equal to 0. According to Eq.~\eqref{eq:expected_hessian}, the principal curvature, $\bar{\kappa}^{\alpha,\beta}$, associated with the expected, dimension-reduced Hessian $H_{\alpha,\beta}$ is also equal to 0, erroneously indicating an apparently flat loss landscape if one would use $\bar{\kappa}^{\alpha,\beta}$ as the main measure of curvature. To compare the convergence of different curvature measures as a function of the number of loss projections $S$, we will now study the ensemble mean
\begin{equation}
\langle X \rangle=\frac{1}{S}\sum_{k=1}^S X^{(k)}
\label{eq:ensemble_average}
\end{equation}
of different quantities of interest $X$ such as Hessian elements and principal curvature measures in dimension-reduced space. Here, $X^{(k)}$ is the $k$-th realization (or sample) of $X$.
\begin{figure}
    \centering
    \includegraphics[width=\textwidth]{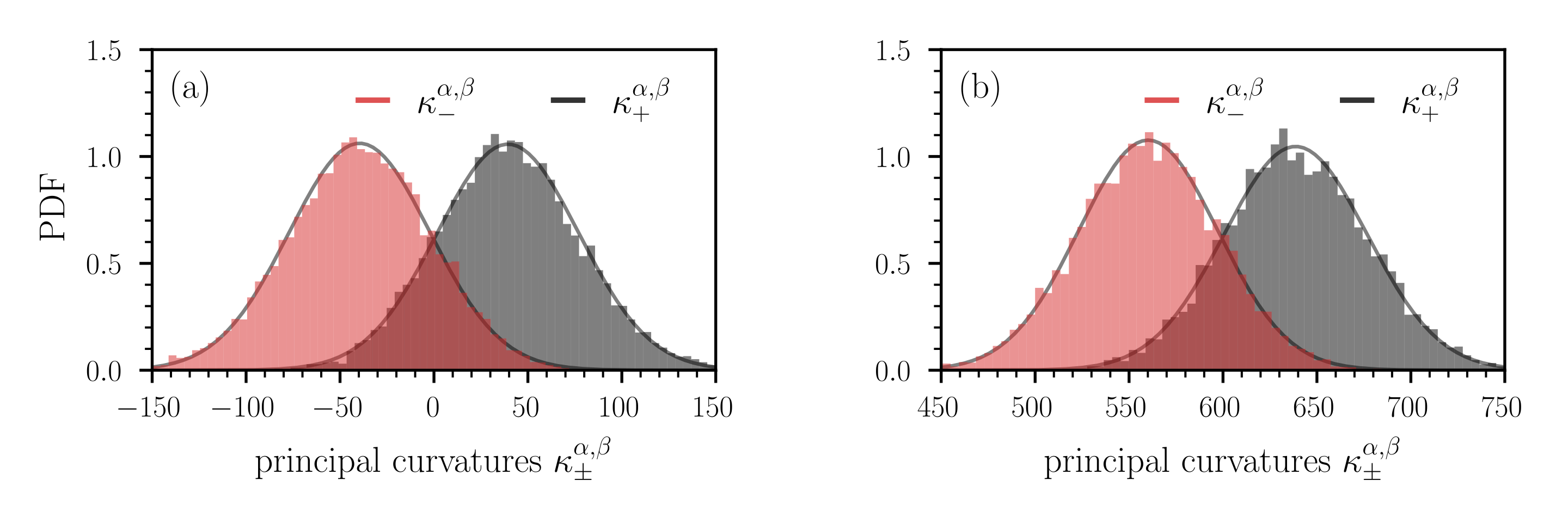}
    \caption{Distribution of principal curvatures $\kappa_{-}^{\alpha,\beta}$ (red bars) and $\kappa_{+}^{\alpha,\beta}$ (black bars). In panels (a) and (b), the loss functions are given by Eqs.~\eqref{eq:loss_symmetric} and \eqref{eq:loss_asymmetric}, respectively. We evaluate the corresponding Hessians \eqref{eq:emp_hessian_1} and \eqref{eq:emp_hessian_2} at the saddle point $\theta^*=(\theta^*_1,\dots,\theta^*_{2n},\theta^*_{2n+1})=(0,\dots,0,1)$. In both loss functions, we set $n=500$ and in loss function~\eqref{eq:loss_asymmetric} we set $\tilde{n}=800$. While in panel (a), the probability $\Pr(\kappa_+^{\alpha,\beta}\kappa_-^{\alpha,\beta}>0)$ that the critical point in the lower-dimensional, random projection does not appear as a saddle is about 0.3, it is 1 in panel (b). Histograms are based on 10,000 random projections that are used to compute $\kappa_{\pm}^{\alpha,\beta}$. Solid grey lines indicate Gaussian approximations of the empirical distributions.}
    \label{fig:curvature_pdfs}
\end{figure}

We first study the dependence of ensemble means $\langle (H_{\alpha,\beta})_{ij} \rangle$ ($i,j\in\{1,2\}$) of elements of the dimension-reduced Hessian $H_{\alpha,\beta}$ on the number of samples $S$. According to Eq.~\eqref{eq:expected_hessian}, the diagonal elements of $\mathds{E}[H_{\alpha,\beta}]$ are proportional to the mean curvature of the original high-dimensional space and are thus useful to examine curvature properties of high-dimensional loss functions. Figure~\ref{fig:hessians_random}(a) shows the convergence of the ensemble means $\langle (H_{\alpha,\beta})_{ij} \rangle$ towards the expected values $\mathds{E}[(H_{\alpha,\beta})_{ij}]$ as a function of $S$. Note that $\mathds{E}[(H_{\alpha,\beta})_{ij}]=0$ for all $i,j$. For a few dozen loss projections, the deviations of the ensemble means from the corresponding expected values reach values larger than 20. A relatively large number of loss projections $S$ between $10^3$--$10^4$ is required to keep these deviations at values that are smaller than about 2--4. The solid black and red lines in Figure~\ref{fig:hessians_random}(b), respectively, show the ensemble means $\langle \kappa_\pm^{\alpha,\beta} \rangle$ and the eigenvalues
\begin{equation}
\langle \tilde{\kappa}_\pm^{\alpha,\beta} \rangle = \frac{1}{2}\left(\langle A\rangle+\langle C\rangle \pm \sqrt{4 \langle B\rangle ^2+(\langle A\rangle-\langle C\rangle)^2}\right)
\label{eq:kappa_tilde_sample}
\end{equation}
of the ensemble-averaged dimension-reduced Hessian as a function of $S$. Since $\mathds{E}[A]=\mathds{E}[C]={(H_\theta)^i}_i$ and $\mathds{E}[B]=0$ [see Eq.~\eqref{eq:expected_hessian}], we have that $\langle \tilde{\kappa}_\pm^{\alpha,\beta} \rangle=\bar{\kappa}^{\alpha,\beta}$ in the limit $S\rightarrow\infty$. In the current example, the ensemble means $\langle \tilde{\kappa}_\pm^{\alpha,\beta} \rangle$ thus approach $\bar{\kappa}^{\alpha,\beta}=0$ for large numbers of samples $S$, represented by the dashed red lines in Figure~\ref{fig:hessians_random}(b). The ensemble means $\langle \kappa_{\pm}^{\alpha,\beta}\rangle $ converge towards values of opposite sign, indicating a saddle point.

For a sample size of $S=10^4$, we show the distribution of the principal curvatures $\kappa_\pm^{\alpha,\beta}$ in Figure~\ref{fig:curvature_pdfs}(a). We observe that the distributions are plausibly Gaussian. We also calculate the probability $\Pr(\kappa_+^{\alpha,\beta}\kappa_-^{\alpha,\beta}>0)$ that the critical point in the lower-dimensional, random projection does not appear as a saddle (\ie, $\kappa_+^{\alpha,\beta}\kappa_-^{\alpha,\beta}>0$). For the example shown in Figure~\ref{fig:curvature_pdfs}(a), we find that $\Pr(\kappa_+^{\alpha,\beta}\kappa_-^{\alpha,\beta}>0)\approx 0.3$. That is, in about 30\% of the simulated projections, the lower-dimensional loss landscape wrongly indicates that it does not correspond to a saddle.

Our first example, which is based on the loss function \eqref{eq:loss_symmetric}, shows that the principal curvatures in the lower-dimensional representation of $L(\theta)$ may capture the saddle behavior in the original space if one computes ensemble means $\langle \kappa_{\pm}^{\alpha,\beta}\rangle$ in the lower-dimensional space [Figure~\ref{fig:hessians_random}(b)]. However, if one first calculates ensemble means of the elements of the dimension-reduced Hessian $H_{\alpha,\beta}$ to infer $\langle \tilde{\kappa}_\pm^{\alpha,\beta} \rangle$, the loss landscape appears to be flat in this example. We thus conclude that different ways of computing ensemble means (either before or after calculating the principal curvatures) may lead to different results with respect to the ``flatness'' of a dimension-reduced loss landscape. 

In the next example, we will show that random projections cannot identify certain saddle points regardless of the underlying averaging process. We consider the loss function
\begin{equation}
L(\theta)=\frac{1}{2}\theta_{2n+1}\left(\sum_{i=1}^{\tilde{n}}\theta_i^2-\sum_{i=\tilde{n}+1}^{2n}\theta_{i}^2\right)\,,\quad n\in\mathbb{Z}_{+},n<\tilde{n}\leq 2 n\,,
\label{eq:loss_asymmetric}
\end{equation}
where we use the convention $\sum_{i=a}^{b} (\cdot) = 0$ if $a>b$. The Hessian at the critical point $(\theta^*_1,\dots,\theta^*_{2n},\theta^*_{2n+1})=(0,\dots,0,1)$ is 
\begin{equation}
H_\theta=\mathrm{diag}(\underbrace{1,\dots,1}_{\tilde{n}~\mathrm{times}},\underbrace{-1,\dots,-1}_{2n-\tilde{n}~\mathrm{times}},0)\,.
\label{eq:emp_hessian_2}
\end{equation}
As in the previous example, the critical point is again a saddle, but the mean curvature is ${H=2(\tilde{n}-n)/N>0}$. In the following numerical experiments, we set $n=500$ and $\tilde{n}=800$. Figure~\ref{fig:hessians_random}(c) shows that the ensemble means $\langle (H_{\alpha,\beta})_{ij}\rangle$ converge towards the expected value $\mathds{E}[(H_{\alpha,\beta})_{ij}]$ as the number of samples increases. We again observe that a relatively large number of random loss projections $S$ between $10^3$ and $10^4$ is required to keep the deviations of ensemble means from their corresponding expected values small. Because of the dominance of positive principal curvatures $\kappa_i^\theta$ in the original space, the corresponding ensemble means of principal curvatures (\ie, $\langle \kappa_{\pm}^{\alpha,\beta} \rangle$, $\langle \tilde{\kappa}_\pm^{\alpha,\beta}\rangle$) in the lower-dimensional representation approach positive values [Figure~\ref{fig:hessians_random}(d)]. The distribution of $\kappa_\pm^{\alpha,\beta}$ indicates that the probability of observing a saddle in the lower-dimensional loss landscape is vanishingly small [Figure~\ref{fig:curvature_pdfs}(b)]. In this second example, both ways of computing ensemble means, before and after calculating the lower-dimensional principal curvatures, mistakenly suggest that the saddle in the original space is a minimum in dimension-reduced space.
\begin{figure}
    \centering
    \includegraphics[width=\textwidth]{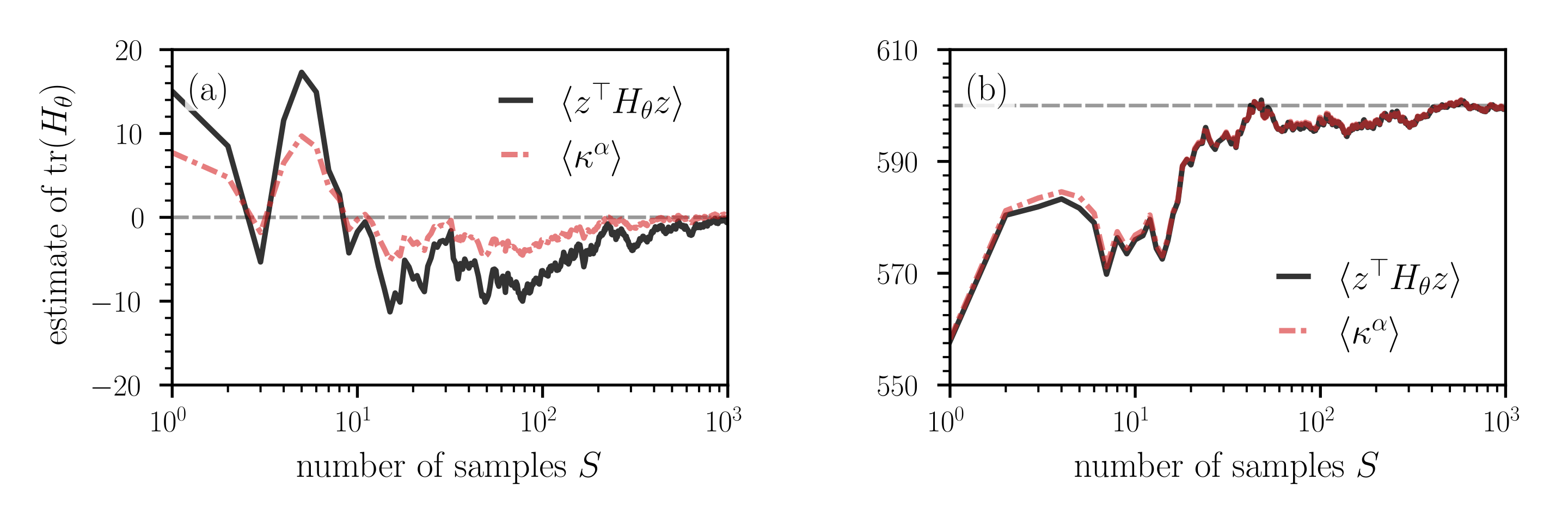}
    \caption{Estimating the trace of the Hessian $H_\theta$. In panels (a) and (b), the loss functions are given by Eqs.~\eqref{eq:loss_symmetric} and \eqref{eq:loss_asymmetric}, respectively. We evaluate the corresponding Hessians \eqref{eq:emp_hessian_1} and \eqref{eq:emp_hessian_2} at the saddle point $\theta^*=(\theta^*_1,\dots,\theta^*_{2n},\theta^*_{2n+1})=(0,\dots,0,1)$. In both loss functions, we set $n=500$ and in loss function~\eqref{eq:loss_asymmetric} we set $\tilde{n}=800$. Solid black and dash-dotted red lines represent Hutchinson [$\langle z^\top H_\theta z \rangle$; see Eq.~\eqref{eq:hutchinson}] and curvature-based ($\langle \kappa^\alpha\rangle$) estimates of $\mathrm{tr}(H_\theta)$, respectively. We compute ensemble means $\langle \cdot \rangle$ as defined in Eq.~\eqref{eq:ensemble_average} for different numbers of random projections $S$. The trace estimates in panels (a) and (b), respectively, converge towards the true trace values $\mathrm{tr}(H_\theta)=0$ and $\mathrm{tr}(H_\theta)=600$ that are indicated by dashed grey lines. In both methods, the same random vectors with elements that are distributed according to a standard normal distribution are used. For the curvature-based estimation of $\mathrm{tr}(H_\theta)$, we perform least-square fits of $L(\theta^*+\alpha\eta)$ over an interval $\alpha\in[-0.05,0.05]$.}
    \label{fig:trace_estimation}
\end{figure}

To summarize, for both loss functions \eqref{eq:loss_symmetric} and \eqref{eq:loss_asymmetric}, the saddle point $\theta^*=(0,\dots,0,1)$ in the original loss function $L(\theta)$ is often misrepresented in lower-dimensional representations $L(\theta+\alpha \eta+\beta\delta)$ if random directions are used. Depending on (i) the employed curvature measure and (ii) the index of the underlying Hessian $H_\theta$ in the original space, the saddle $\theta^*=(0,\dots,0,1)$ often appears erroneously as either a minimum, maximum, or an almost flat region.

If the critical point were a minimum or maximum (\ie, a critical point associated with a positive definite or negative definite Hessian $H_\theta$), it would be correctly represented in the corresponding expected random projection because the sign of its principal curvature $\bar{\kappa}^{\alpha,\beta}$, which is proportional to the sum of all eigenvalues $\kappa_i^\theta$ of the Hessian $H_{\theta}$ in the original loss space, would be equal to the sign of the principal curvatures $\kappa_i^\theta$. However, such points are scarce in high-dimensional loss spaces~\cite{DBLP:conf/nips/DauphinPGCGB14,baldi1989neural}.

Finally, because of the confounding factors associated with the inability of random projections to correctly identify saddle information, it does not appear advisable to use ``flatness'' around a critical point in a lower-dimensional random loss projection as a measure of generalization error~\cite{DBLP:conf/nips/Li0TSG18}. 
\subsection{Hessian trace}
\label{sec:hessian_trace}
We now use the loss functions~\eqref{eq:loss_symmetric} and \eqref{eq:loss_asymmetric} to compare the convergence behavior between the original Hutchinson method~\eqref{eq:hutchinson} and the curvature-based trace estimation~\eqref{eq:trace_estimate}. Instead of two random directions, we use one random Gaussian direction $\eta$ and perform quadratic least-square fits for 50 equidistant values of $\alpha$ in the interval $[-0.05,0.05]$ to extract estimates of $\mathrm{tr}(H_\theta)$ from $L(\theta^*+\alpha\eta)$. We use the same random Gaussian directions in Hutchinson's method.

Figure~\ref{fig:trace_estimation} shows how the Hutchinson and curvature-based trace estimates converge towards the true trace values, 0 for the loss function~\eqref{eq:loss_symmetric} and 600 for the loss function \eqref{eq:loss_asymmetric} with $n=500$ and $\tilde{n}=800$. Given that we use the same random vectors in both methods, their convergence behavior towards the true trace value is similar. With the curvature-based method, one can produce Hutchinson-type trace estimates without computing Hessian-vector products. However, it requires the user to specify an appropriate interval for the quantity $\alpha$ so that the mean curvature can be properly estimated in a quadratic-approximation regime. It also requires a sufficiently large number of points in this interval. Therefore, it may be less accurate than the original Hutchinson method.
\subsection{Hessian directions}
\label{sec:hessian_directions}
\begin{figure}
    \centering
    \includegraphics[width=\textwidth]{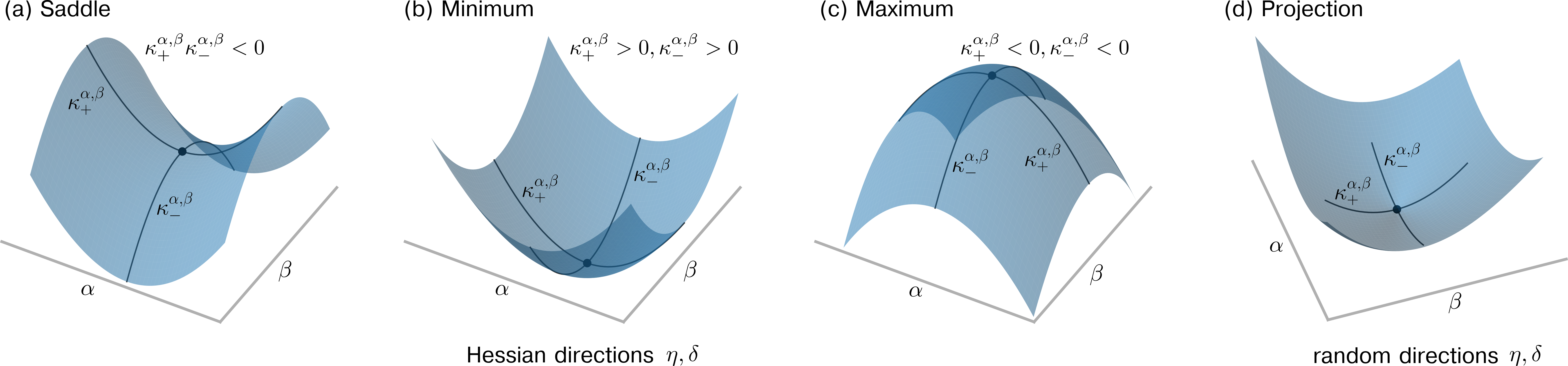}
    \caption{Dimensionality-reduced loss $L(\theta^*+\alpha \eta+\beta\delta)$ of Eq.~\eqref{eq:loss_asymmetric} with $n=900,\tilde{n}=1000$ for different directions $\eta,\delta$. (a--c) The directions $\eta,\delta$ correspond to eigenvectors of the Hessian $H_\theta$ of Eq.~\eqref{eq:loss_asymmetric}. If the eigenvalues associated with $\eta,\delta$ have different signs, the corresponding loss landscape is a saddle as depicted in panel (a). If the eigenvalues associated with $\eta,\delta$ have the same sign, the corresponding loss landscape is either a minimum (both signs are positive) as shown in panel (b) or a maximum (both signs are negative) as shown in panel (c). Because there is an excess of $\tilde{n}-n=100$ positive eigenvalues in $H_\theta$, a projection onto a dimension-reduced space that is spanned by the random directions $\eta,\delta$ is often associated with an apparently convex loss landscape. An example of such an apparent minimum is shown in panel (d). We selected a single pair of random directions $\eta,\delta$ (\ie, no averaging over random directions has been performed).}
    \label{fig:loss_projection}
\end{figure}

Given the described shortcomings of random projections (\eg, in correctly identifying saddles of high-dimensional loss functions), we suggest to use Hessian directions (\ie, the eigenbasis of $H_\theta$) as directions $\eta,\delta$ in $L(\theta^*+\alpha \eta + \beta \delta)$. For Eq.~\eqref{eq:loss_asymmetric} with $n=900,\tilde{n}=1000$, we show projections along different Hessian directions in Figure~\ref{fig:loss_projection}(a--c). We observe that different Hessian directions indicate different types of critical points in dimension-reduced space. If the eigenvalues associated with the Hessian directions $\eta,\delta$ have different signs, the corresponding lower-dimensional loss landscape is a saddle [Figure~\ref{fig:loss_projection}(a)]. If both eigenvalues have the same sign, the loss landscape is either a minimum [Figure~\ref{fig:loss_projection}(b): both signs are positive] or it is a maximum [Figure~\ref{fig:loss_projection}(c): both signs are negative]. If one uses a random projection instead, the resulting lower-dimensional loss landscape often appears to be a minimum in this example [Figure~\ref{fig:loss_projection}(d)]. To quantify the proportion of random projections that correctly identify the saddle with $n=900,\tilde{n}=1000$, we generated 10,000 realizations of $\kappa_{\pm}^{\alpha,\beta}$ [see Eq.~\eqref{eq:kapp_alpha_beta}]. We find that the signs of $\kappa_{\pm}^{\alpha,\beta}$ were different in only about 0.5\% of all simulated realizations. That is, in this example the principal curvatures $\kappa_{\pm}^{\alpha,\beta}$ indicate a saddle in only about 0.5\% of the studied projections. 

In the next section, we will show that Hessian directions that are associated with the largest-magnitude positive and negative eigenvalues of $H_\theta$ are useful to appropriately identify saddle information and visually study the geometric properties of saddle points in high-dimensional loss spaces of image classifiers.
\section{Applications to neural networks}
\label{sec:appl_nn}
\begin{figure}
    \centering
    \includegraphics[width=\textwidth]{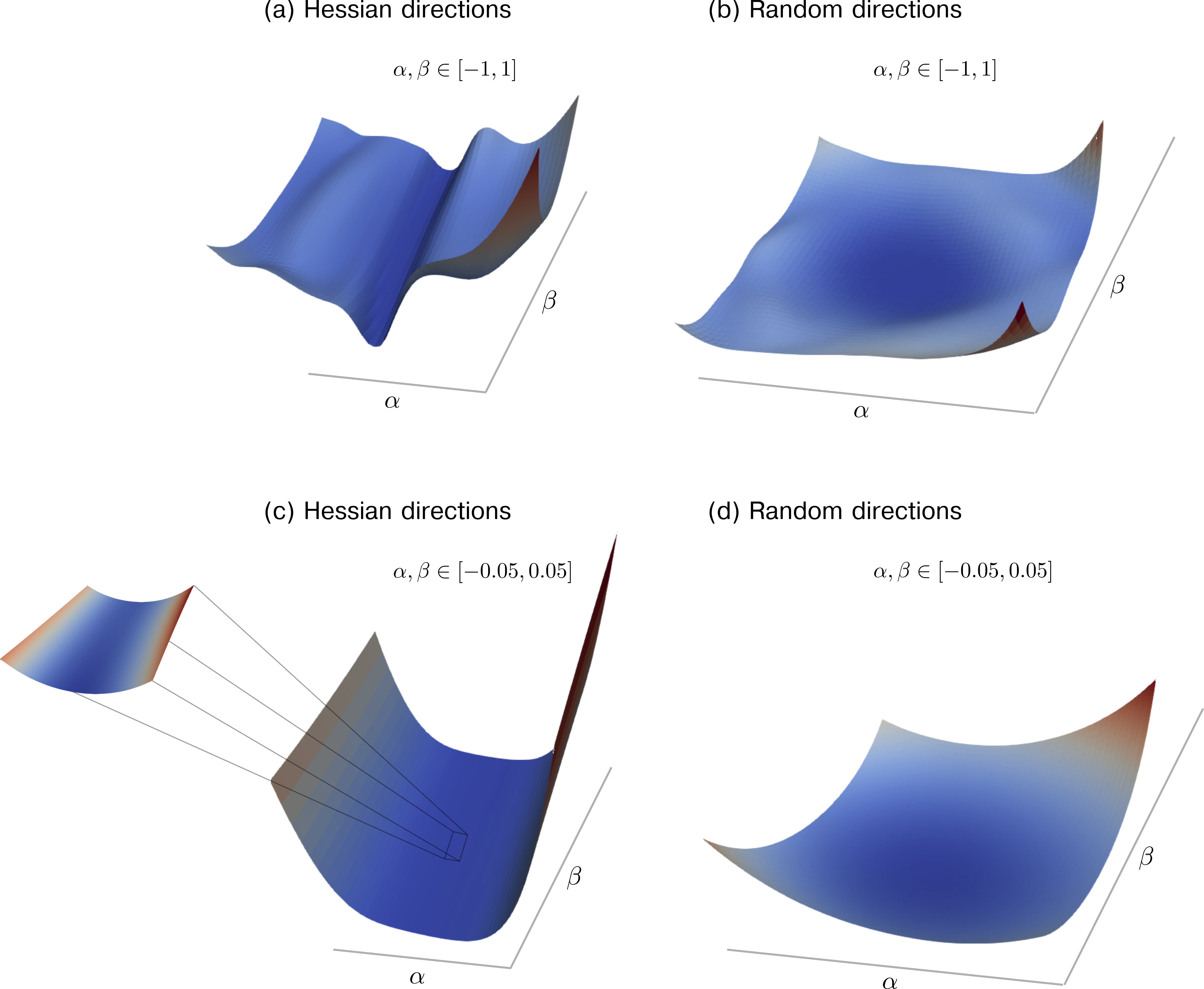}
    \caption{Loss landscape projections for ResNet-56. (a,c) The projection directions $\eta,\delta$ are given by the eigenvectors associated with the largest and smallest eigenvalues of the Hessian $H_\theta$, respectively. The zoomed inset in panel (c) shows the loss landscape for $(\alpha,\beta)\in[-0.01,0.005]\times[-0.05,0.05]$. We observe a decreasing loss along the negative $\beta$-axis. (b,d) The projection directions $\eta,\delta$ are given by random vectors. We selected a single pair of random directions $\eta,\delta$ (\ie, no averaging over random directions has been performed). The domains of $(\alpha,\beta)$ in panels (a,b) and (c,d) are $[-0.05,0.05]\times[-0.05,0.05]$ and $[-1,1]\times[-1,1]$, respectively. All shown cross-entropy loss landscapes are based on evaluating the CIFAR-10 training dataset that consists of 50,000 images.}
    \label{fig:resnet_56}
\end{figure}
We now compare projections $L(\theta^*+\alpha \eta+\beta \delta)$ using (i) {\em dominant Hessian directions} (\ie, eigenvectors that are associated with the largest-magnitude positive and negative eigenvalues of $H_\theta$) and (ii) random directions. Hessian directions were computed using HVPs without an explicit representation of $H_\theta$ (see Algorithm~\ref{alg:hessian_directions}). We first compute the largest-magnitude eigenvalue and then use an annihilation method~\cite{burden2015numerical} to compute the second largest-magnitude eigenvalue of opposite sign. More details on the annihilation algorithm are provided in Appendix~\ref{app:annihilation}. Other deflation techniques can be used to find additional Hessian directions.

Instead of explicitly representing the Hessian $H_\theta$, Algorithm~\ref{alg:hessian_directions} evaluates products between $H_\theta$ and a vector $v\in\mathbb{R}^N$ using the identity
\begin{equation}
\nabla_\theta [(\nabla_\theta L)^\top v]=(\nabla_\theta \nabla_\theta L) v +(\nabla_\theta L)^\top \nabla_\theta v= H_\theta v\,.
\end{equation}
In the first step, the gradient $(\nabla_\theta L)^\top$ is computed using reverse-mode autodifferentiation (AD) to then compute the scalar product $[(\nabla_\theta L)^\top v]$. In the the second step, we again apply reverse-mode AD to the computational graph associated with the scalar product $[(\nabla_\theta L)^\top v]$. Because the vector $v$ does not depend on $\theta$ (\ie, $\nabla_\theta v=0$), the result is $H_\theta v$. One may also use forward-mode AD in the second step to provide a more memory efficient implementation.\footnote{Different forward-mode AD functions have been made available in \texttt{PyTorch} since at least version 1.11, starting from March 2022.}

We used the \texttt{LinearOperator} module that is available in the \texttt{Python} package \texttt{scipy} to represent HVPs. As an alternative to implementing HVPs manually, one may use the \texttt{hvp} function provided in the \texttt{torch.autograd.functional} module. Moreover, the \texttt{Python} package \texttt{PyHessian}~\cite{yao2020pyhessian} can be also used to evaluate Hessians of high-dimensional functions in a distributed manner. For the computation of dominant Hessian directions, we used the \texttt{scipy} function \texttt{eigsh} that is based on the implicitly restarted Lanczos method~\cite{lehoucq1998arpack}. 

The computational cost of Algorithm~\ref{alg:hessian_directions} can be summarized as follows. For the HVPs, the computational cost is $\mathcal{O}(N)$~\cite{pearlmutter1994fast}, where $N$ is the number of neural-network parameters. To identify extremal eigenvalues along with their corresponding eigenvectors using a Lanczos-type method, one has to perform a certain number of matrix-vector multiplications. The required number of iterations depends on several factors, including the desired accuracy, the characteristics of the underlying Hessian matrix, and the choice of the initial vector.

We first focus on a ResNet-56 architecture that has been trained in \cite{DBLP:conf/nips/Li0TSG18} on {CIFAR-10} using stochastic gradient descent (SGD) with Nesterov momentum. The number of parameters is 855,770.  The training and test losses at the local optimum found by SGD are ${9.20\times 10^{-4}}$ and 0.29, respectively; the corresponding accuracies are 100.00 and 93.66, respectively. 

\begin{algorithm}[H]
\caption{Compute dominant Hessian directions}
\label{alg:hessian_directions}
\begin{algorithmic}[1]
\STATE{$L_1=\mathrm{LinearOperator((N,N),~matvec=hvp)}$}\COMMENT{initialize linear operator for HVP calculation}
\STATE{eigval1, eigvec1 = solve\_lm\_evp($L_1$)\COMMENT{compute largest-magnitude eigenvalue and corresponding eigenvector associated with operator $L_1$}}
\STATE{shifted\_hvp(vec) = hvp(vec) - eigval1*vec}\COMMENT{define shifted HVP}
\STATE{$L_2=\mathrm{LinearOperator((N,N),~matvec=shifted\_hvp)}$}\COMMENT{initialize linear operator for shifted HVP calculation}
\STATE{eigval2, eigvec2 = solve\_lm\_evp($L_2$)}\COMMENT{compute largest-magnitude eigenvalue and corresponding eigenvector associated with operator $L_2$}
\STATE{eigval2 += eigval1}
\IF{eigval1 $>=$ 0}
\STATE{maxeigval, maxeigvec = eigval1, eigvec1}
\STATE{mineigval, mineigvec = eigval2, eigvec2}
\ELSE
\STATE{maxeigval, maxeigvec = eigval2, eigvec2}
\STATE{mineigval, mineigvec = eigval1, eigvec1}
\ENDIF
\RETURN maxeigval, maxeigvec, mineigval, mineigvec
\end{algorithmic}
\end{algorithm}

In accordance with \cite{DBLP:conf/nips/Li0TSG18}, we apply filter normalization to random directions. This and related normalization methods are often employed when generating random projections. One reason is that simply adding random vectors to parameters of a neural network loss function (or parameters of other functions) does not consider the range of parameters associated with different elements of that function. As a result, random perturbations may be too small or large to properly resolve the influence of certain parameters on a given function.

When calculating Hessian directions, we are directly taking into account the parameterization of the underlying functions that we want to visualize. Therefore, there is no need for an additional rescaling of different parts of the perturbation vector. Still, reparameterizations of a neural network can result in changes of curvature properties (see, \eg, Theorem 4 in \cite{DBLP:conf/icml/DinhPBB17}).

Figure~\ref{fig:resnet_56} shows the two-dimensional projections of the loss function (cross entropy loss) around a local critical point. The smallest and largest eigenvalues are $-16.4$ and $5007.9$, respectively. This means that the found critical point is a saddle with a maximum negative curvature that is more than two orders of magnitude smaller than the maximum positive curvature at that point. The saddle point is clearly visible in Figure~\ref{fig:resnet_56}(a,c). We observe in the zoomed inset in Figure~\ref{fig:resnet_56}(c) that the loss decreases along the negative $\beta$-axis.

If random directions are used, the corresponding projections indicate that the optimizer converged to a local minimum and not to a saddle point [Figure~\ref{fig:resnet_56}(b,d)]. Overall, the ResNet-56 visualizations that we show in Figure~\ref{fig:resnet_56} exhibit structural similarities to those that we generated using the simple loss model \eqref{eq:loss_asymmetric} [Figure~\ref{fig:loss_projection}(d)].

\begin{figure}
    \centering
    \includegraphics[width=0.8\textwidth]{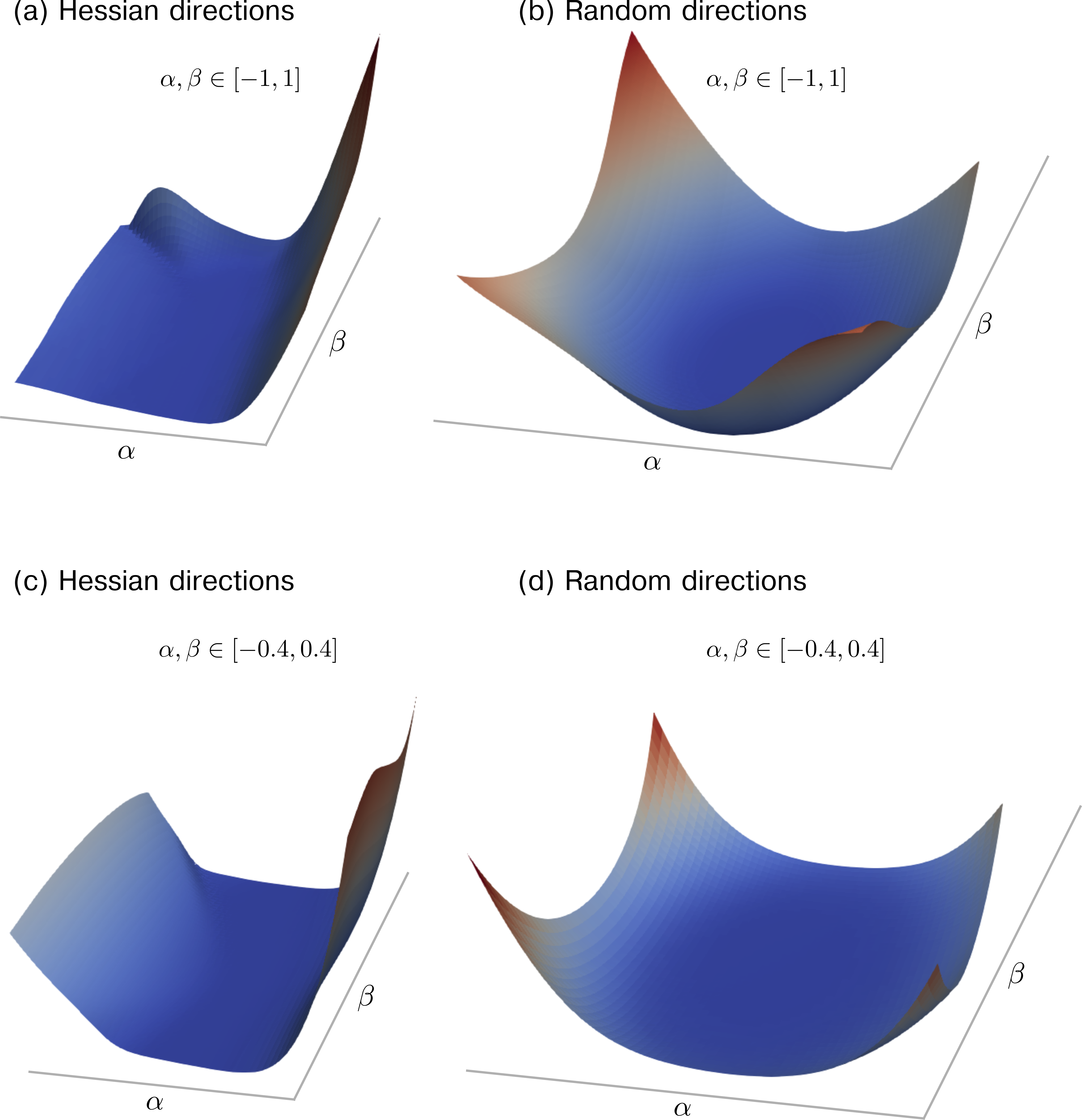}
    \caption{Loss landscape projections for DenseNet-121. (a,c) The projection directions $\eta,\delta$ are given by the eigenvectors associated with the largest and smallest eigenvalues of the Hessian $H_\theta$, respectively. (b,d) The projection directions $\eta,\delta$ are given by random vectors. We selected a single pair of random directions $\eta,\delta$ (\ie, no averaging over random directions has been performed). The domains of $(\alpha,\beta)$ in panels (a,b) and (c,d) are $[-0.4,0.4]\times[-0.4,0.4]$ and $[-1,1]\times[-1,1]$, respectively. All shown cross-entropy loss landscapes are based on evaluating the CIFAR-10 training dataset that consists of 50,000 images.}
    \label{fig:densenet_121}
\end{figure}

As a second neural-network structure, we consider DenseNet-121 that has also been trained in \cite{DBLP:conf/nips/Li0TSG18} on CIFAR-10 using SGD with Nesterov momentum. In this neural network, the number of parameters is 6,956,298. Figure~\ref{fig:densenet_121} shows the projections $L(\theta^*+\alpha \eta+\beta \delta)$ of the loss function (cross entropy loss) around a local optimum. The training and test losses of the local optimum found by SGD are ${8.07\times 10^{-4}}$ and $1.69\times 10^{-1}$, respectively; the corresponding accuracies are 100.00 and 95.63, respectively. We again observe the typical saddle shape of $L$ if dominant Hessian directions are used and an apparent local minimum for projections along random directions. One should keep in mind that the principal curvatures in the random direction plot are given by $\kappa_{\pm}^{\alpha,\beta}$ [see Eq.~\eqref{eq:kapp_alpha_beta}]. At a critical point, the principal curvatures in dimension-reduced, random-projection space are, on average, equal to the sum of the principal curvatures in the original space. If only very few, small-magnitude negative principal curvatures are present in the original space, the expected random projections are associated with an apparent local minimum in dimension-reduced, random-projection space.

Analyzing the Hessian $H_\theta$ of DenseNet-121 around its local optimum, we find that the minimum and maximum principal curvatures are -109.1 and 1937.5, respectively. In comparison to ResNet-56, the relative difference between these two curvatures is substantially smaller. In particular the largest-magnitude negative principal curvature in DenseNet-121 is by a factor of about 7 smaller than that in ResNet-56, indicating that more substantial loss improvements may be possible in the former by changing the neural-network parameters $\theta$ along the negative curvature direction.
\begin{figure}
    \centering
    \includegraphics[width=0.9\textwidth]{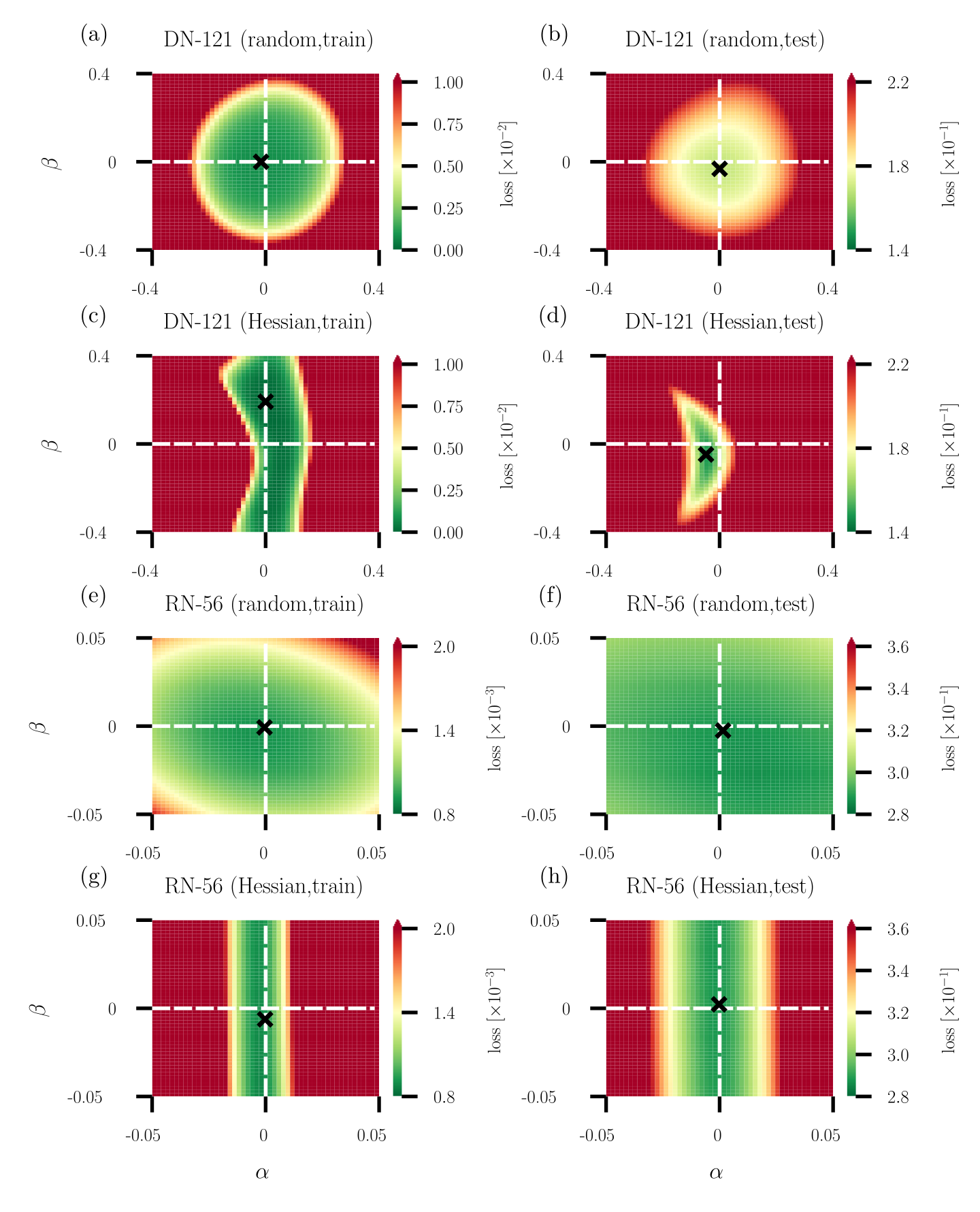}
    \caption{Heatmaps of cross-entropy loss functions along random and dominant Hessian directions. (a--c) Loss heatmaps for DenseNet-121 [(a,c): training data, (c,d): test data]. (e--h) Loss heatmaps for ResNet-56 [(e,g): training data, (f,h): test data]. Random directions are used in panels (a,b,e,f) while Hessian directions are used in panels (c,d,g,h). Green and red regions indicate small and large loss values, respectively. Black crosses indicate the positions of loss minima in the shown domain.}
    \label{fig:train_test}
\end{figure}

In Figure~\ref{fig:train_test}, we compare changes in the projected loss along random and dominant Hessian directions for DenseNet-121 and ResNet-56. In addition to evaluating the loss on training data (50,000 images) as in Figures~\ref{fig:resnet_56} and \ref{fig:densenet_121}, we also provide a comparison to loss changes in the test dataset (10,000 images). Black crosses in Figure~\ref{fig:train_test} indicate loss minima. For both neural networks, we observe that random projections are associated with loss values that increase along both directions $\delta,\eta$ and for both training and test data [Figure~\ref{fig:train_test}(a,b,e,f)]. For these projections, the loss minima are very close to the origin of the loss space. Different observations can be made if Hessian directions are employed in the loss projections. Figure~\ref{fig:train_test}(c) shows that the training loss minimum of DenseNet-121 in the Hessian projection is not located at the origin but at $(\alpha,\beta)\approx(0,0.2)$. A value of $\beta\approx0.2$ means that one has to move along the negative curvature direction to find a loss function value that is smaller than the one at the origin. We find that the value of the loss at that point is more than one order of magnitude smaller than the original loss at the origin or the smallest loss found in a random-projection plot. Similar observations can be made in the test dataset. However, the smallest loss value found in the Hessian direction plot is associated with a relatively large loss in the test dataset [Figure~\ref{fig:train_test}(d)]. Using Hessian directions to improve training and test performance requires one to balance potential performance improvements in the test and training datasets. Because of the relatively small negative principal curvature of ResNet-56, one can only achieve small loss improvements along the corresponding Hessian direction and differences in minimum losses between random and Hessian projections are less pronounced [Figure~\ref{fig:train_test}(e--h)].

In Appendix~\ref{app:cifar}, we study random and Hessian loss projections of three additional architectures, namely ResNet-56 without skip connections~\cite{DBLP:conf/nips/Li0TSG18}, MobileNetV2~\cite{GitLabPyTorch}, and GoogLeNet~\cite{GitLabPyTorch}. For ResNet-56 without skip connections, we find that the training loss minimum in the Hessian projection is associated with slightly larger training and test accuracies than the training loss minimum in the random projection. For MobileNetV2 and GoogLeNet, we find that the minimum training loss in the Hessian projection is associated with an either slightly larger training or test accuracy compared to the training loss minimum in the random projection. In the examples that we study in Appendix~\ref{app:cifar}, the training and test loss minima in the Hessian projections are better aligned than those in the corresponding random projections. Table~\ref{tab:summary_of_results} in Appendix~\ref{app:cifar} summarizes the results for all five architectures.

Image classification architectures and their training are already highly optimized to achieve very good performance in common image classification tasks. Our results show that it is possible to improve training loss values along the dominant negative curvature direction. Smaller loss values may be also achievable with further gradient-based training iterations, so one has consider tradeoffs between gradient and curvature-based optimization methods. Also, assessing overall performance of dominant Hessian directions on this task is not straightforward. For instance, loss measurements and accuracy measurements are not proportional to one another for this image classification task (see Table~\ref{tab:summary_of_results} in Appendix~\ref{app:cifar}). A further complication is that, since these architectures are highly optimized and the underlying generative processes are unknown, it is difficult to interpret variations in performance that may appear between training loss, test loss, and accuracy. Therefore, to better address the question of whether better performance in training by dominant Hessian directions can translate to better test performance, we need to control for some of the suspected confounds that appear in the image classification task. So, we also study possible loss improvements along the dominant negative curvature direction in a function approximation task~\cite{adcock2021gap} that we describe in Appendix~\ref{app:function_approx}. 

Unlike image classification, function approximation involves sampling from a known generative distribution. Our results, summarized in  Appendix~\ref{app:function_approx}, show that both training loss and test loss are smaller in our functional approximation task for Hessian projections than for random projections. Specifically, for 2-layer and 10-layer fully connected neural networks (FCNNs), with 20,501 and 101,301 parameters, respectively, training loss is more than 9\% smaller for the 2-layer network and 26\% smaller for the 10-layer network for Hessian projections than for random projections and test loss is 1\% smaller and 6\% smaller, respectively. Using this function approximation example, we have also included in the Supplemental Information~\cite{video} an animation that shows the evolution of random and Hessian loss projections during training.
\section{Discussion and conclusion}
\label{sec:conclusion}
Given that a global understanding of high-dimensional loss landscapes remains still very limited~\cite{cooper2021global}, one- and two-dimensional projections of such functions are a common tool to study the geometric properties of loss regions nearby critical points and along optimizer trajectories~\cite{DBLP:journals/corr/GoodfellowV14,DBLP:conf/nips/Li0TSG18,DBLP:conf/emnlp/HaoDWX19,wu2020adversarial,DBLP:conf/ida/HoroiHRLWK22}. For an informed way of projecting high-dimensional functions to lower-dimensional representations, it is important to mathematically interpret the geometric information contained in such visualizations. To put it in the words of Ker-Chau Li~\cite{li1992principal}: ``There are too many directions to project a high-dimensional data set and unguided plotting can be time-consuming and fruitless.'' Therefore, it is important to understand how projection directions $\eta,\delta$ affect the resulting loss visualizations. A standard choice in the literature is to use random Gaussian directions to obtain lower-dimensional loss representations $L(\theta^*+\alpha\eta+\beta\delta)$. Combining concepts from high-dimensional probability and differential geometry, we show that saddle points in the original space may not be correctly identified as such in lower-dimensional representations if such random projections are used. 

As formalized in Eq.~\eqref{eq:mean_curvature}, the expected principal curvature $\bar{\kappa}^{\alpha,\beta}=\mathrm{tr}(H_\theta)$, obtained by averaging over Hessian elements ${(H_\theta)^i}_i$ in a dimension-reduced Hessian [see Eq.~\eqref{eq:expected_hessian}], is proportional to the mean curvature $H=\mathrm{tr}(H_\theta)/N$ in the original loss space. The connection between $\bar{\kappa}^{\alpha,\beta}$ and $\mathrm{tr}(H_\theta)$ shows that Hutchinson-type trace estimates can be computed directly from curvature information in lower-dimensional random projections~\cite{hutchinson1989stochastic}.

Depending on the value of the mean curvature in the original loss space, saddle points appear, on average, as either minima ($H>0$), maxima ($H<0$), or almost flat regions ($H\approx 0$). Instead of computing ensemble means of elements of dimension-reduced Hessians, one may also compute ensemble means of principal curvatures. As illustrated in different numerical experiments, both ways of averaging are associated with different geometric interpretations and random projections are generally not able to correctly identify saddle points. We therefore propose to study projections along dominant Hessian directions that are associated with the largest-magnitude positive and negative principal curvatures in the original space. Similar methods have been developed and applied in related work on exploratory data analysis~\cite{li1992principal}. Thus, our work may be viewed as part of a recent  ``Hessian turn'' in machine learning characterized by computationally tractable methods for unlocking Hessian information in non-linear and high-dimensional settings \cite{yao2020pyhessian,liao2021hessian}. Projecting high-dimensional loss functions along Hessian directions provides a genuine way of illustrating their geometric properties around critical points. Negative curvature directions may be used as a starting point for further improving the performance of a given neural network. Analyzing principal curvatures (or eigenvalues of $H_\theta$) at critical points $\theta^*$ is also useful to identify global minima. It has been shown by Cooper~\cite{cooper2021global} that for any global minimum of $L(\theta)$, the corresponding Hessian $H_\theta$ has $r n$ positive
eigenvalues, $N- rn$ eigenvalues equal to 0, and no negative eigenvalues. Here, $n$ and $r$ are the number of input samples (\ie, data points) and the output dimension, respectively.

Our numerical experiments on different image classifiers and a function approximation task are in accordance with our reasoning and identify marked differences in the projections that are based on Hessian and random directions. Our results also address the ongoing debate on the connection between flatness of a certain loss region and generalizability (\ie, the ability to effectively use trained neural networks on unseen data). Dinh \etal~\cite{DBLP:conf/icml/DinhPBB17} provided a detailed discussion on different mechanisms (\eg, reparameterization of flat regions) that indicate that sharp minima can achieve good generalization properties. Their work also highlighted the importance of clear definitions of flatness. Our work shows that flatness in randomly projected loss landscapes might be just apparent and a consequence of (large) principal curvatures of opposite sign that cancel out each other.

There are many interesting and worthwhile directions for future research building on the work reported here. To provide more insights into the geometric properties of high-dimensional loss functions of deep neural networks, future work may study other convolutional architectures~\cite{DBLP:conf/bmvc/ZagoruykoK16,DBLP:conf/cvpr/MartinelFM19}, vision transformers~\cite{DBLP:conf/iclr/DosovitskiyB0WZ21}, and residual multi-layer perceptrons~\cite{touvron2022resmlp}. Our results suggest that negative curvature directions may be useful to find better local optima, but further research is needed to gain a clearer understanding of the tradeoffs associated with optimization protocols that are based on gradients and curvature information. More research on learning algorithms that minimize loss functions along negative curvature directions~\cite{alain2019negative} can contribute to the development of more effective optimizers. Such guided optimization algorithms may be of particular use in applications (\eg, neural network\textendash based control~\cite{asikis2022neural,bottcher2022ai,bottcher2023gradientfree} and function approximation in numerical analysis~\cite{adcock2021gap}) where training data can be generated with a high resolution and used to approximate a function as closely as possible. In applications where one wishes to reduce generalization error, the proposed mean-curvature estimation method may be also useful to improve existing curvature regularization approaches~\cite{moosavi2019robustness}. Furthermore, to improve our understanding of the properties of loss landscapes, it may be helpful to devote more attention to the study of connections between random matrix theory and the characterization of high-dimensional functions~\cite{bray2007statistics,baron2022eigenvalue}. For example, for critical points of Gaussian fields in high-dimensional spaces, Bray and Dean~\cite{bray2007statistics} showed that there is a monotonic relation between the index of the Hessian (\ie, the number of negative eigenvalues) and the loss value: the larger the index, the larger the corresponding loss. Bray and Dean~\cite{bray2007statistics} also showed that the eigenvalues of the Hessian at a critical point of such fields are distributed according to Wigner's semicircle law~\cite{wigner1951statistical} up to an additional shift. However, traditional random-matrix-theory results such as Wigner's semicircle law or the Mar\v cenko--Pastur law~\cite{marvcenko1967distribution} are based on various simplifying assumptions, including Gaussian weight, error, and data distributions, which do not hold in practical applications of neural networks~\cite{liao2021hessian}. 
\section*{Acknowledgements}
We thank Mingtao Xia and Thomas Asikis for helpful discussions and an anonymous reviewer for their valuable feedback and suggestions.
\section*{Funding}
LB acknowledges funding from the ARO through grant W911NF-23-1-0129.
\bibliography{refs,1_master}
\appendix
\section{Concentration inequality}
\label{app:concentration}
For $\eta_i,\delta_i\sim\mathcal{N}(0,1)$ and $X_i=\eta_i+\delta_i,Y_i=\eta_i-\delta_i$, our goal is to construct a concentration inequality for $|N^{-1} \sum_{i} \eta_i \delta_i|=|(2 N)^{-1} \sum_i X_i^2-1+1-Y_i^2|=|(\tilde{X}-1+1-\tilde{Y})/2|$, where $\tilde{X}=N^{-1}\sum_{i=1}^N X_i^2$ and $\tilde{Y}=N^{-1}\sum_{i=1}^N Y_i^2$. Notice that $\tilde{X}$ and $\tilde{Y}$ are independent because the covariance associated with the corresponding bivariate normal distribution, which is valid in the limit of large $N$, vanishes. Using that the cumulative distribution function of the sum of independent random variables is given by the convolution of the individual distributions, we obtain
\begin{align}
\begin{split}
\Pr\left(\frac{1}{N}\sum_{i=1}^N \eta_i \delta_i \geq \epsilon \right)&=\Pr\left(\frac{1}{2N}\sum_{i=1}^N X_i^2-1+1-Y_i^2 \geq \epsilon \right)\\
&=\Pr\left(\tilde{X}-1+1-\tilde{Y} \geq 2\epsilon \right)\\
&=\int_0^\infty \Pr\left(\tilde{X}-1\geq 2\epsilon+\tilde{Y}-1|\tilde{Y}=y\right)f_{\tilde{Y}}(y)\,\mathrm{d}y\\
&\leq \int_0^\infty e^{-\frac{N (2\epsilon+y-1)^2}{4}}f_{\tilde{Y}}(y)\,\mathrm{d}y\,,
\end{split}
\end{align}
where we used inequality \eqref{eq:concentration_bound_11} for $\epsilon\rightarrow 0$ in the last step. In the limit of large $N$, the distribution function $f_{\tilde{Y}}(y)$ of the random variable $\tilde{Y}$ approaches a Gaussian with mean $1$ and variance $2/N$. With this approximation, we find
\begin{equation}
\Pr\left(\frac{1}{N}\sum_{i=1}^N \eta_i \delta_i \geq \epsilon \right)\leq \frac{e^{-{N \epsilon^2 \over 2}}}{\sqrt{2}}\,.
\end{equation}
The same bound applies to $\Pr\left(1-\tilde{X}+\tilde{Y}-1 \geq 2\epsilon \right)$, so we have
\begin{equation}
\Pr\left(\left|\frac{1}{N}\sum_{i=1}^N \eta_i \delta_i\right| \geq \epsilon \right)\leq \sqrt{2} e^{-{N \epsilon^2 \over 2}}\,.
\end{equation}

For two independent random variables $X,Y$, a similar bound can be derived using the inequality
\begin{align}
\begin{split}
\Pr\left(X+Y\geq \epsilon\right)&=\int_{\mathbb{R}^2}\mathds{1}_{x+y\geq \epsilon} f_{X,Y}(x,y)\,\mathrm{d}x\mathrm{d}y\\
&=\int_{\mathbb{R}^2}\mathds{1}_{x+y\geq \epsilon} \mathds{1}_{x\geq \epsilon/2} f_{X,Y}(x,y)\,\mathrm{d}x\mathrm{d}y\\
&+\int_{\mathbb{R}^2}\mathds{1}_{x+y\geq \epsilon} \mathds{1}_{x< \epsilon/2} f_{X,Y}(x,y)\,\mathrm{d}x\mathrm{d}y\\
&\leq\int_{-\infty}^\infty \mathds{1}_{x\geq\epsilon/2} f_X(x)\,\mathrm{d}x+\int_{-\infty}^\infty \mathds{1}_{y\geq\epsilon/2} f_Y(y)\,\mathrm{d}y\\
&=\Pr\left(X\geq \epsilon/2\right)+\Pr\left(Y\geq \epsilon/2\right)\,,
\end{split}
\label{eq:inequality_sumXY}
\end{align}
where $f_{X,Y}(x,y)=f_X(x) f_Y(y)$ denotes the joint distribution associated with random variables $X,Y$. Using inequality \eqref{eq:inequality_sumXY}, we obtain
\begin{equation}
\Pr\left(\left|\frac{1}{N}\sum_{i=1}^N \eta_i \delta_i\right| \geq \epsilon \right)\leq 4 e^{-{N \epsilon^2 \over 4}}\,.
\end{equation}
\section{Annihilation method}
\label{app:annihilation}
Let $A\in\mathbb{R}^{N\times N}$ be a symmetric matrix with eigenvalues $\{\lambda_i\}_{i=1,\dots,N}$ satisfying
\begin{equation}
|\lambda_1|>|\lambda_2|\geq\dots\geq|\lambda_N|\quad\text{and}\quad|\lambda_k-\lambda_1|>|\lambda_j-\lambda_1|\quad\text{for~$j\neq k$}\,.
\label{eq:eigenvalues}
\end{equation}
Here, $\lambda_k$ is the largest-magnitude eigenvalue that belongs to the set of eigenvalues of opposite sign of $\lambda_1$. An example of a matrix with eigenvalues that satisfy Eq.~\eqref{eq:eigenvalues} is a Hessian $H_\theta$ at a saddle point with distinct largest-magnitude eigenvalues (\ie, principal curvatures). Because the matrix $A$ is symmetric, it has an orthonormal eigenbasis which we denote by $\{v_j\}_{j=1,\dots,N}$. Furthermore, let ${B=A-\lambda_1 \mathds{1}_{N\times N}}$. For a vector $z_0\in \mathbb{R}^N$, we calculate
\begin{align}
z_1\equiv B z_0=Az_0-\lambda_1 z_0=\sum_{j=1}^N c_j A v_j -\lambda_1 \sum_{j=1}^N c_j v_j = \sum_{j=2}^N c_j (\lambda_j-\lambda_1) v_j\,,
\end{align}
where $c_j=\langle v_j, z_0\rangle$. The corresponding Rayleigh quotient associated with $z_n=B z_{n-1}$ ($n\in\{1,2,\dots\}$) is
\begin{equation}
\lambda^{(n)}\equiv \frac{z_n^\top B z_n}{z_n^\top z_n}=\frac{\sum_{j=2}^N c_j^{2} (\lambda_j-\lambda_1)^{2n+1}}{\sum_{j=2}^N c_j^{2} (\lambda_j-\lambda_1)^{2n}}\,,
\end{equation}
where we used that $\langle v_i, v_j\rangle=0$ if $i\neq j$ and 1 otherwise. Notice that $\lambda^{(n)}\rightarrow \lambda_k-\lambda_1$ in the limit $n\rightarrow\infty$. The described method is reminiscent of eigenvalue annihilation as commonly used in deflation techniques~\cite{burden2015numerical}. The main difference of the outlined deflation method w.r.t.\ standard annihilation approaches is that we are not interested in the second largest-magnitude eigenvalue, but in the largest-magnitude eigenvalue that belongs to the set of eigenvalues of opposite sign of $\lambda_1$. The eigenvalues $\lambda_1,\lambda_k$ can be identified with the largest-magnitude negative and positive principal curvatures and the corresponding eigenvectors are the dominant Hessian directions.
\section{Image classification}
\label{app:cifar}
\begin{figure}
    \centering
    \includegraphics[width=0.75\textwidth]{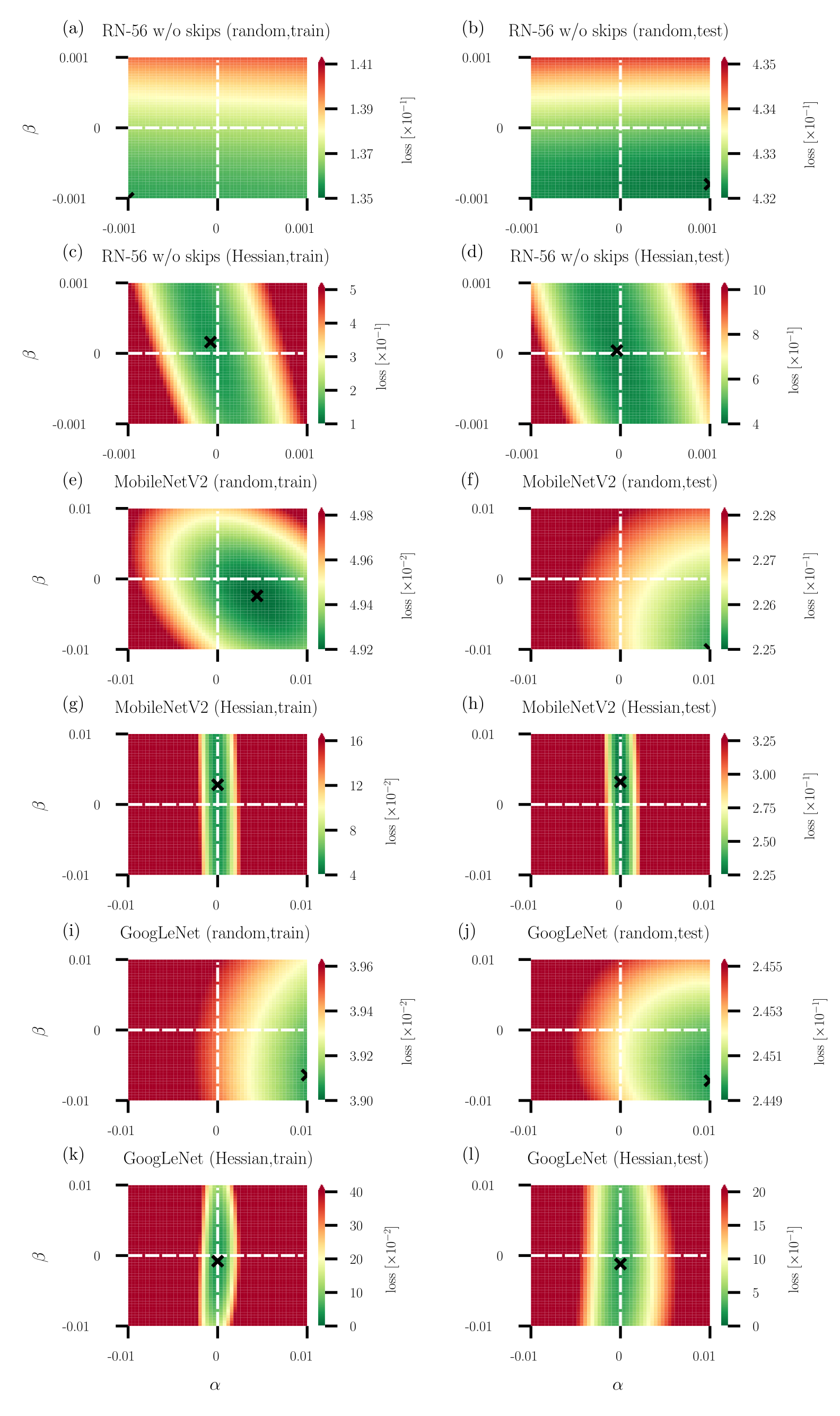}
    \caption{Heatmaps of cross-entropy loss functions along random and dominant Hessian directions. (a--c) Loss heatmaps for ResNet-56 without skip connections [(a,c): training data, (c,d): test data]. (e--h) Loss heatmaps for MobileNetV2 [(e,g): training data, (f,h): test data]. (i--l) Loss heatmaps for GoogLeNet [(i,j): training data, (k,l): test data]. Random directions are used in panels (a,b,e,f,i,j) while Hessian directions are used in panels (c,d,g,h,k,l). Green and red regions indicate small and large loss values, respectively. Black crosses indicate the positions of loss minima in the shown domain.}
    \label{fig:cifar_2_heatmaps}
\end{figure}
To complement our results on loss landscape visualizations of common image classifiers (see Section~\ref{sec:appl_nn}), we consider three additional architectures, namely the ResNet-56 architecture from the main text but without skip connections~\cite{DBLP:conf/nips/Li0TSG18}, MobileNetV2~\cite{GitLabPyTorch}, and GoogLeNet~\cite{GitLabPyTorch}. The numbers of parameters of these three networks are 853,018, 2,236,682, and 5,490,986, respectively. As in the main text, all architectures were trained using SGD with Nesterov momentum. Further implementation details are available in \cite{DBLP:conf/nips/Li0TSG18,GitLabPyTorch}. 

A complete summary of results for all five image classifiers is given in Table~\ref{tab:summary_of_results}. For ResNet-56 without skip connections, MobileNetV2, and GoogLeNet, we show heatmaps of the corresponding cross-entropy loss along random and dominant Hessian directions in Figure~\ref{fig:cifar_2_heatmaps}. Overall, Table~\ref{tab:summary_of_results} shows that random projections and Hessian directions are comparable in performance, as measured by minimum loss and maximum accuracy. However, the heatmaps in Figure~\ref{fig:cifar_2_heatmaps} reveal differences between the two methods: first, in terms of the loss landscape around a learned minimizer; second, and more importantly, in terms of the degree of alignment between the location of the optimal training parameters and the location of the optimal test parameters.  

%\begin{landscape}
\begin{table}
\centering
{\scriptsize
\renewcommand*{\arraystretch}{1.5}
\begin{tabular}{l cc cc c}\toprule
        & Train loss & Test loss          & Train acc & Test acc  & dHd \\ \hline
RN-56    & $9.20 \times 10^{-4}$ & $\mathbf{2.88\times 10^{-1}}$  & $\mathbf{99.998}$   & 93.66   & $\{-16.4, 5007.9\}$\\
 \, \, \, rand proj & $9.10\times 10^{-4}$ & $\mathbf{2.88\times 10^{-1}}$ & $\mathbf{99.998}$ & $\mathbf{93.68}$ & --\\
  \, \, \, Hess & $\mathbf{8.75\times 10^{-4}}$ & $2.89\times 10^{-1}$ & $\mathbf{99.998}$ & 93.66 & --\\ \hline
DN-121   & $8.07 \times 10^{-4}$ & $\mathbf{1.69\times 10^{-1}}$  & $\mathbf{100}$ & 95.63   & $\{-109.1, 1937.5\}$\\
 \, \, \, rand proj & $8.03\times 10^{-4}$ & $\mathbf{1.69\times 10^{-1}}$ & $\mathbf{100}$ & $\mathbf{95.64}$ & --\\
  \, \, \, Hess & $\mathbf{6.14\times 10^{-5}}$ & $2.46\times 10^{-1} $ & $\mathbf{100}$ & 95.32 & --\\ \hline
RN-56 (w/o skips)  & $1.37\times 10^{-1}$ &  $4.33\times 10^{-1}$  & 95.546 & 86.69   & $\{-39308.9,50064044.0\}$ \\
 \, \, \, rand proj & $1.36\times 10^{-1}$ & $\mathbf{4.32\times 10^{-1}}$ & 95.578     & 86.73   & --\\
  \, \, \, Hess & $\mathbf{1.35\times 10^{-1}}$ & $4.33\times 10^{-1}$ & $\mathbf{95.584}$ & $\mathbf{86.74}$ & --  \\ \hline
MobileNetV2  & $4.93\times 10^{-2}$ & $2.27\times 10^{-1}$ & 99.518 &  93.91  & $\{-405.0,1061115.0\}$ \\
 \, \, \, rand proj  & $4.92\times 10^{-2}$ & $\mathbf{2.26\times 10^{-1}}$     & $\mathbf{99.524}$    &  93.90  & --\\
  \, \, \, Hess & $\mathbf{4.90\times 10^{-2}}$         & $\mathbf{2.26\times 10^{-1}}$ &  99.516  & $\mathbf{93.93}$  & --  \\ \hline 
GoogLeNet  &       $3.95\times 10^{-2}$          &  $2.45\times 10^{-1}$  &   99.878    &  92.82   &   $\{-8208.8,1427483.0\}$\\
 \, \, \, rand proj & $\mathbf{3.91\times 10^{-2}}$   & $2.45\times 10^{-1}$ & 99.878     & $\mathbf{92.89}$   & --\\
  \, \, \, Hess     & $\mathbf{3.91\times 10^{-2}}$   &  $\mathbf{2.44\times 10^{-1}}$       & $\mathbf{99.884}$  & 92.76 & -- \\ \bottomrule
\end{tabular}
}
\vspace{1mm}
\caption{\textbf{Summary of image classification results.} We summarize the four performance measures training loss, test loss, training accuracy, and test accuracy of ResNet-56, DenseNet-121, ResNet-56 without skip connections, MobileNetV2, and GoogLeNet. For each neural network, we report all performance measures after initial training in the first row of each cell. The second and third rows in each cell list the performance measures that are associated with minimum training loss in random projections (``rand proj'') and Hessian projections (``Hess''), respectively. The column ``dHd'' lists the principal curvatures associated with the dominant Hessian directions at the initial training optimum.}
\label{tab:summary_of_results}
\end{table}
%\end{landscape}

The lack of skip connections in the studied ResNet-56 architecture is associated with larger loss values (training loss: $1.37\times 10^{-1}$; test loss: $4.33\times 10^{-1}$) and smaller accuracies (training accuracy: 95.55; test accuracy: 86.69) compared to the ResNet-56 loss values reported in the main text. Because of the large absolute values between $10^4-10^8$ of the principal curvatures associated with the dominant Hessian directions, we study loss projections in a small domain $[-10^{-3},10^{-3}]\times[-10^{-3},10^{-3}]$ around the found optimum. Figure~\ref{fig:cifar_2_heatmaps}(a,b) shows that the minimum training and test loss points in the random projection heatmaps are not aligned. The minimum training loss is $1.36\times 10^{-1}$ with an associated accuracy of 95.58. The corresponding test loss and accuracy are $4.32\times 10^{-1}$ and 86.73, respectively. In the Hessian direction heatmaps that we show in Figure~\ref{fig:cifar_2_heatmaps}(c,d), the minimum training loss is $1.35\times 10^{-1}$, slightly smaller than the minimum loss in the random-projection plot. The corresponding accuracies are slightly larger than the corresponding random projection values (training accuracy: 95.584 vs.\ 95.578; test accuracy: 86.74 vs.\ 86.73).

Also for MobileNetV2 we observe in Figure~\ref{fig:cifar_2_heatmaps}(e,f) that the minimum training and test loss values in the random projection heatmaps are not aligned. The minimum training loss is $4.92\times 10^{-2}$ and the corresponding test loss is $2.26\times 10^{-1}$. The associated training and test accuracies are 99.52 and 93.90, respectively.
The minimum training and test loss points in the Hessian projections in Figure~\ref{fig:cifar_2_heatmaps}(g,h) are more closely aligned than in the random projections. The minimum training loss in the Hessian projection is $4.90\times 10^{-2}$, again slightly smaller than in the random projection. The training accuracy is also slightly smaller (99.516 vs.\ 99.524) while the corresponding test accuracy is slightly larger (93.93 vs.\ 93.90) compared to the random projection values.

In contrast to the other random projections in  Figure~\ref{fig:cifar_2_heatmaps}, the GoogLeNet random projection in Figure~\ref{fig:cifar_2_heatmaps}(i,j) has better aligned training and test loss minima. The minimum training loss is $3.91\times 10^{-2}$ and the corresponding test loss is $2.45\times 10^{-1}$. The associated training and test accuracies are 99.88 and 92.89. The minimum training loss in the Hessian projection is also $3.91\times 10^{-2}$, but the training accuracy is slightly larger than in the corresponding random projection (99.884 vs.\ 99.878). The corresponding test accuracy is slightly smaller (92.76
 vs.\ 92.89) than in the random projection.
\section{Function approximation}
\label{app:function_approx}
\begin{figure}
    \centering
    \includegraphics[width=0.66\textwidth]{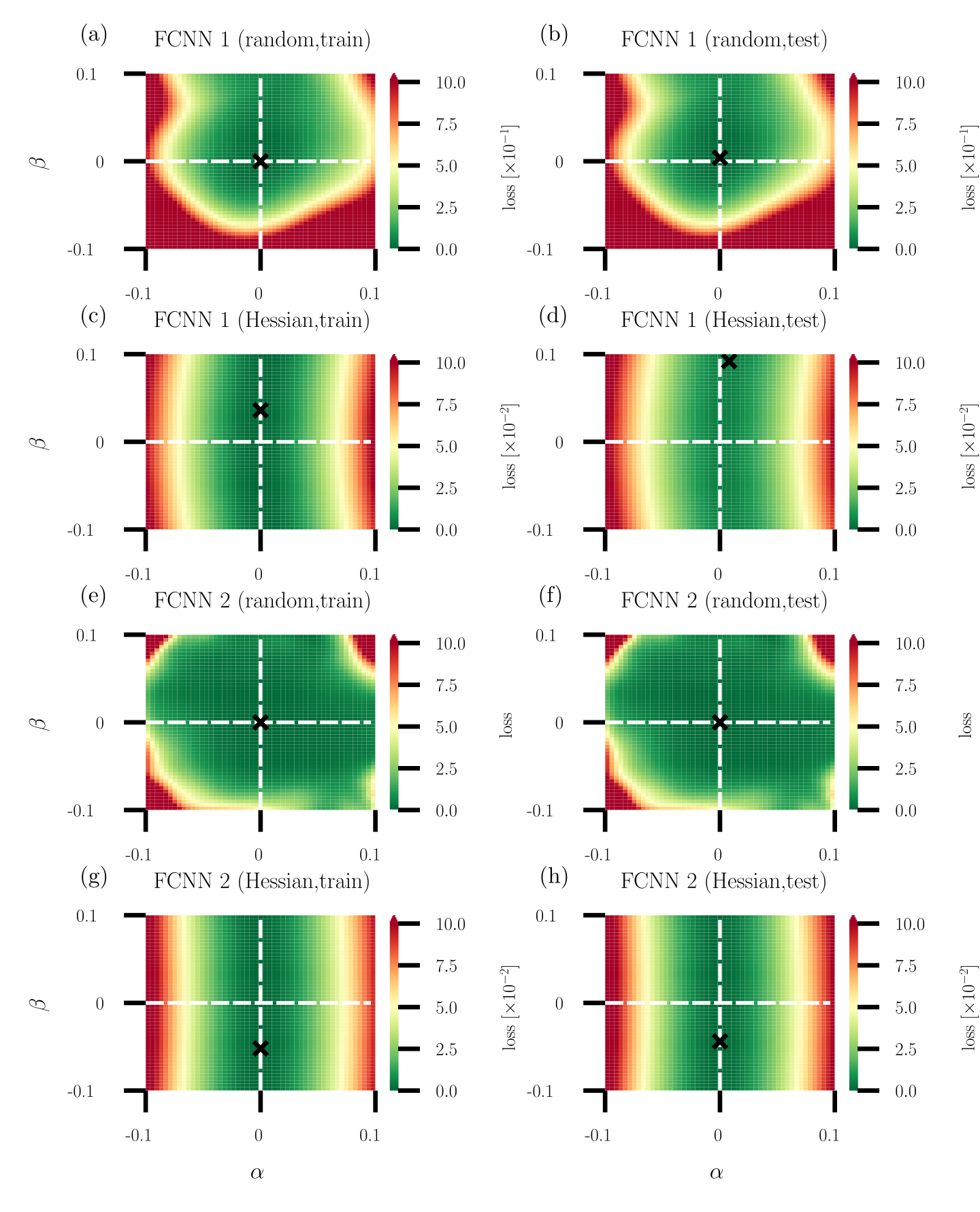}
    \caption{Heatmaps of the mean squared error (MSE) loss along random and dominant Hessian directions for a function approximation task. (a--c) Loss heatmaps for a fully connected neural network (FCNN) with 2 layers and 100 ReLU activations per layer [(a,c): training data, (c,d): test data]. (e--h) Loss heatmaps for an FCNN with 10 layers and 100 ReLU activations per layer [(e,g): training data, (f,h): test data]. Random directions are used in panels (a,b,e,f) while Hessian directions are used in panels (c,d,g,h). Green and red regions indicate small and large mean squared error (MSE) loss values, respectively. Black crosses indicate the positions of loss minima in the shown domain. Both neural networks, FCNN 1 and FCNN 2, are trained to approximate the smooth one-dimensional function \eqref{eq:func_approx}.}
    \label{fig:train_test_function_approximation}
\end{figure}
We compare loss function visualizations that are based on random and Hessian directions in a function approximation task. In accordance with \cite{adcock2021gap}, we consider the smooth one-dimensional function
\begin{equation}
    f(x)=\log(\sin(10x)+2)+\sin(x)\,,
\label{eq:func_approx}
\end{equation}
where $x\in [-1,1)$. To approximate $f(x)$, we use two fully connected neural networks (FCNNs) with 2 and 10 layers, respectively. Each layer has 100 ReLU activations. The numbers of parameters of the 2 and 10 layer architectures are 20,501 and 101,301, respectively. The training data is based on 50 points that are sampled uniformly at random from the interval $[-1,1)$. We train both neural networks using a mean squared error (MSE) loss function and SGD with a learning rate of 0.1. The 2 and 10 layer architectures are respectively trained for 100,000 and 50,000 epochs to reach loss values of less than $10^{-4}$. The best model was saved and evaluated by calculating the MSE loss for 1,000 points that were sampled uniformly at random from the interval $[-1,1)$.

Figure~\ref{fig:train_test_function_approximation} shows heatmaps of the training and test loss landscapes of both neural networks along random and dominant Hessian directions. Black crosses in Figure~\ref{fig:train_test_function_approximation} indicate loss minima. As in the main text, we observe for both neural networks that random projections are associated with loss values that increase along both directions $\delta,\eta$ and for both training and test data [Figure~\ref{fig:train_test_function_approximation}(a,b,e,f)]. For these projections, we find that the loss minima are very close to or at the origin of the loss space. The situation is different in the loss projections that are based on Hessian directions. Figure~\ref{fig:densenet_121}(c) shows that the loss minimum for the 2-layer FCNN is not located at the origin but at $(\alpha,\beta)\approx(0,0.04)$. We find that the value of the training loss at that point is more than 9\% smaller than the smallest training loss found in a random-projection plot. The corresponding test loss is about 1\% smaller than the test loss associated with the smallest training loss in the random-projection plot. For the 10-layer FCNN, the smallest training loss in the Hessian direction projection is more than 26\% smaller than the smallest training loss in the randomly projected loss landscape [Figure~\ref{fig:train_test_function_approximation}(e,g)]. The corresponding test loss in the Hessian direction plot is about 6\% smaller than the corresponding test loss minimum in the random direction plot [Figure~\ref{fig:train_test_function_approximation}(f,h)]. Notice that both the training and test loss minima in the random direction heatmaps in Figure~\ref{fig:train_test_function_approximation}(e,f) are located at the origin while they are located at $(\alpha,\beta)\approx(0,-0.05)$ in the Hessian direction heatmaps in Figure~\ref{fig:train_test_function_approximation}(g,h).

In the Supplemental Information~\cite{video}, we include an animation that shows the evolution of lower-dimensional loss projections of the 10-layer FCNN.
\end{document}